\pdfoutput=1
\documentclass[11pt]{article}

\usepackage[]{ACL2023}
\usepackage{times}
\usepackage{latexsym}

\usepackage[T1]{fontenc}
\usepackage[utf8]{inputenc}

\usepackage{microtype}

\usepackage{inconsolata}

\usepackage{graphicx}
\usepackage{booktabs}
\usepackage{amsmath}
\usepackage{pifont}% http://ctan.org/pkg/pifont
\usepackage{multirow}
\newcommand{\cmark}{\ding{51}}%
\newcommand{\xmark}{\ding{55}}%
\usepackage{array,multirow}
\usepackage{array}
\newcolumntype{C}[1]{>{\centering\arraybackslash}p{#1}}
\usepackage{graphicx}
\usepackage{subfig}

\usepackage{soul}

\newcommand*{\affaddr}[1]{#1} % No op here. Customize it for different styles.
\newcommand*{\affmark}[1][*]{\textsuperscript{\rm #1}}
\newcommand*{\email}[1]{\texttt{#1}}

\title{Zero-Shot Text Classification via Self-Supervised Tuning}

\author{
Chaoqun Liu\thanks{$^{*}$Chaoqun Liu  and Guizhen Chen are under the Joint PhD Program between Alibaba and Nanyang Technological University. }~~\affmark[12]\;
Wenxuan Zhang\thanks{$^{\dag}$Wenxuan Zhang is the corresponding author.}~~\affmark[2]\; 
Guizhen Chen$^{*}$\affmark[12]\;
Xiaobao Wu\affmark[1]\; \\ 
\textbf{Anh Tuan Luu\affmark[1]\;
Chip Hong Chang\affmark[1]\;
Lidong Bing\affmark[2]}\\
\affaddr{\affmark[1]Nanyang Technological University, Singapore}, 
\affaddr{\affmark[2]DAMO Academy, Alibaba Group}\\
\email{\{chaoqun.liu,guizhen.chen,saike.zwx,l.bing\}@alibaba-inc.com}\\
\email{\{xiaobao002,echchang,anhtuan.luu\}@ntu.edu.sg}\\
}

\begin{document}
\maketitle

\begin{abstract}
% Existing solutions to zero-shot text classifications are either sensitive to the choices of patterns and verbalizers or relying on large-scale annotated data to  fine-tune pre-trained language models. 
% \wxb{Mention the problem of this kind of method.} 
Existing solutions to zero-shot text classification either conduct prompting with pre-trained language models, which is sensitive to the choices of templates, or rely on large-scale annotated data of relevant tasks for meta-tuning. In this work, we propose a new paradigm based on self-supervised learning to solve zero-shot text classification tasks by tuning the language models with unlabeled data, called self-supervised tuning. By exploring the inherent structure of free texts, we propose a new learning objective called first sentence prediction to bridge the gap between unlabeled data and text classification tasks. After tuning the model to learn to predict the first sentence in a paragraph based on the rest, the model is able to conduct zero-shot inference on unseen tasks such as topic classification and sentiment analysis. Experimental results show that our model outperforms the state-of-the-art baselines on 7 out of 10 tasks. Moreover, the analysis reveals that our model is less sensitive to the prompt design. Our code and pre-trained models are publicly available at \url{https://github.com/DAMO-NLP-SG/SSTuning}.
% Without the need for further training \isak{this is unnecessary since zero-shot already gives ppl impression that training is no need}, 
 %Since the model sees an extensive number of labels during tuning, it is more robust and has better transferability.
\end{abstract}

% \begin{abstract}
% Previous works fine-tune pre-trained language models on a large number of labeled datasets to achieve good zero-shot performance on text classification tasks. However, such a paradigm is sub-optimal due to limited training sample size and label space. To overcome the limitations, we propose a self-supervised learning method to fine-tune the language model on unlabeled datasets. After tuning the model to learn to predict the first sentence in a paragraph, the model obtains the capability to do zero-shot inference on text classification tasks like topic classification and sentiment analysis. Without additional training, our model outperforms the state-of-the-art baselines on several tasks. Our code and checkpoints will be publicly available. %Since the model sees an extensive number of labels during tuning, it is more robust and has better transferability.
% \isak{remember to revise the abs accordingly at last}

% \end{abstract}

\section{Introduction}

%%%%%%%%%%%%%%%%%%%%%%%%%%%%%%%%%%%%%%%%%%%%%%%%%%%%%%%%%%%%%%%%%%%%%%%%%%%%%
Recent advances in pre-trained language models (PLMs) have brought enormous performance improvements in a large variety of NLP tasks \citep{GPT1,naacl19/bert}.
% These paradigm shifts towards leveraging generic features learnt by existing PLMs for different related tasks are motivated by the high data cost required for learning each new NLP task afresh due to the nuances of language features. 
These paradigm shifts towards leveraging generic features learnt by 
% existing 
PLMs are driven by the high data cost required for learning each new NLP task afresh.
% \isak{delete above sentence}
One promising learning method that echoes this paradigm shift is 
% This development also enables the possibility of 
zero-shot text classification, which predicts text labels on unseen tasks.
% As labeled data for the target domain is no longer an obstacle for downstream tasks, zero-shot setting has attracted plentiful research attention. 
% As labeled data is no longer a necessity for relearning new feature representations for untrained specific tasks, zero-shot text classification has attracted considerable research attention in recent years \citep{iclr22/flan,iclr22/T0,emnlp22/uniMC}. 
Zero-shot text classification has attracted considerable research attention in recent years \citep{iclr22/flan,iclr22/T0,emnlp22/uniMC}, as labeled data is no longer a necessity for relearning new feature representations for untrained specific tasks.
% \wxb{Try to squeeze a line with only one or two words.}

\begin{figure}[t]
    \centering
    \includegraphics[width=\linewidth]{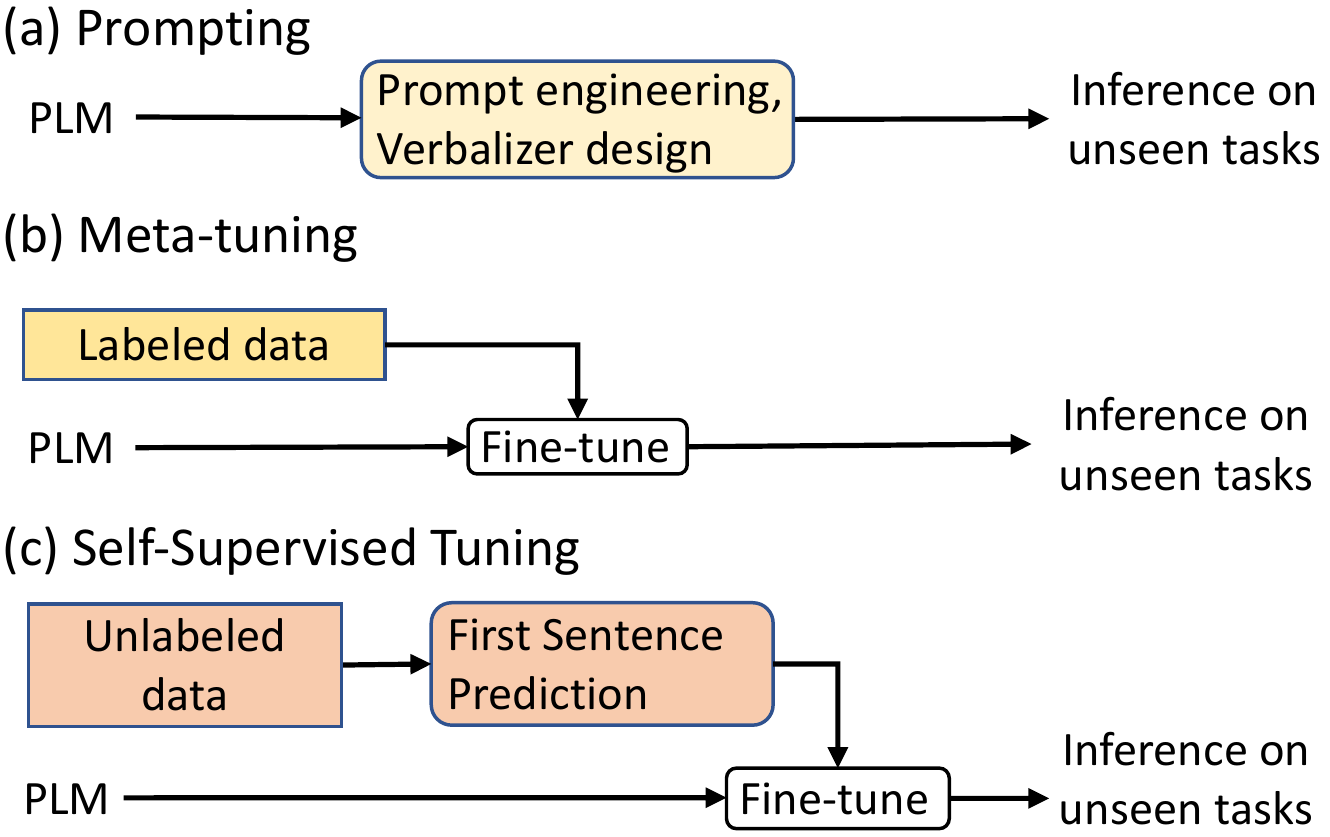}
    \caption{Zero-shot learning approaches: (a) prompting, (b) meta-tuning, and (c) our proposed self-supervised tuning method.} 
    %\wxb{Why the data construction is connected to PLM in (d)?}}
    % \wxb{How about giving the data construction a new name? Like first-sentence data construction.}
    %\isak{better not mention too many models in the figure, making the paradigm comparison clear is enough}\isak{what do you want to convey via this graph?}
    %\isak{too many colors, maybe unnecessary}\cq{changed from 5 colors to 3}
    \label{fig:architecture}
\end{figure}

Existing studies on zero-shot text classification can be briefly classified into two types, as shown in Figure~\ref{fig:architecture}.
The first type is prompting, which uses PLMs to %generate verbalizers \isak{generate?} as 
predict labels with designed templates and verbalizers (Figure~\ref{fig:architecture} (a)). 
This can be achieved by leveraging the generation capability of large language models \citep{npls20/gpt3,PaLM}, or reformulating text classification task as a mask-filling task % in a cloze style \isak{masking-filling and close are the same things} 
\citep{NAACL21/iPET,ECAL21/PET}. 
Likewise, generation-based methods \citep{nlps22/superGen,corr/zerogen} and mining-based methods \citep{emnlp22/mining-based} also rely on prompting to generate or filter noisy labeled samples, which are used for further fine-tuning.
% Prompting can also be used to generate labeled samples for generation-based methods \citep{nlps22/superGen} or filter noisy mined text for mining-based methods \citep{emnlp22/mining-based}. %\isak{this sentence is for?}
%\wxb{Briefly mention some examples.}
The second type is meta-tuning which fine-tunes a PLM on a collection of labeled data of related tasks before conducting inference on unseen tasks  (Figure~\ref{fig:architecture} (b)). By reformulating the annotated data into instruction templates \citep{iclr22/flan,iclr22/T0}, question-answer pairs \citep{emnlp20/unifiedQA,emnlp21/meta-tuning}, multiple-choice questions \citep{emnlp22/uniMC} or entailment pairs \citep{emnlp19/TE,naacl22/TE_TowardsOT,aaai23/TE_multiple_options}, and fine-tuning on them, PLMs perform well on unseen tasks. % \guizhen{\st{with the similar formulation}}.
%\wxb{For example, xxx}.
%
%Weakly-supervised tuning, the last one, fine-tunes a PLM with some generated training samples from unlabeled data. \wxb{For example, xxx}.

Despite the achieved performance, existing methods have several limitations. % for wider applications.
% Specifically,
% \wxb{Give a conclusion sentence about the limitation of prompting here.}
Prompting has shown to be sensitive to the choice of patterns and verbalizers \citep{emnlp22/mining-based}. This makes it difficult to design different templates specifically for each task. In addition, generation-based and mining-based methods require fine-tuning PLMs for each downstream task, which is inefficient for deployment. 
On the other hand, meta-tuning relies on labeled data of relevant tasks or in specific formats to facilitate the learning of desired patterns.
% meta tuning relies on laborious and expensive data labeling.
%Such datasets have a limited number of training samples and limited label space, creating a huge gap between training and zero-shot inference. \isak{delete the above sentence} 
The requirement for such large-scale annotated data narrows its application scope. %\isak{meta tuning relies on labeled data of relevant tasks or in specific formats to equip the model capturing the desired patterns. The requirement for such large-scale annotated data limits its application scope. (You can still discuss the detailed weakness, but may be unnecessary here)}
% \wxb{Add two more sentences to describe the problems of meta tuning.}
%

% % To address the above issues, we in this paper propose a new framework for zero-shot text classification based on self-supervised learning (SSL). 
% % SSL has been widely used in pre-training \cite{naacl19/bert,iclr20/albert} and contrastive learning \citep{corr/survey_ssl}. 
% % To the best our knowledge, this is the first work that applies SSL at tuning stage for zero-shot classification. We call it \textbf{Self-Supervised Tuning} (\textbf{SST}).
% % \isak{
% To address the above issues, we propose to leverage self-supervised learning (SSL) for the zero-shot text classification task. SSL has been widely used during the pre-training stage of PLMs to alleviate the need for large-scale human annotations
% \cite{naacl19/bert,iclr20/albert} by exploiting the intrinsic structure of free texts.
% %, one can obtain effective supervision signals for training models. learning certain modeling abilities, which can be used to directly classify text in a zero-shot manner. \isak{need a transition here} 
% It can also be used to learn better sentence representations in a contrastive manner \citep{tkde/ssl_generative_or_contrastive,emnlp21/simcse}. The benefits for the downstream tasks make it promising for zero-shot learning. \isak{this is weak, somehow echo the weakness you mentioned before}
% To our best knowledge, this is the first work to apply SSL at the tuning stage for zero-shot classification, which we refer to it as self-supervised tuning (SSTuning). 
To address the above issues, we propose to leverage self-supervised learning (SSL) for zero-shot text classification tasks. SSL has been widely used during the pre-training stage of PLMs to alleviate the need for large-scale human annotations
\cite{naacl19/bert,iclr20/albert} by exploiting the intrinsic structure of free texts. Therefore, with a suitable SSL objective, the model is able to capture certain patterns with the auto-constructed training data and can be applied to a wide range of downstream tasks in a zero-shot manner without specific designs. To our best knowledge, this is the first work to exploit SSL at the tuning stage for zero-shot classification, which we refer to as self-supervised tuning (SSTuning). 

% \wxb{Also mention how the proposed method can address the problems of prompting.}
% For text classification tasks, the label can be regarded as the summarization of the text from a certain perspective, like topic and sentiment. We notice that in many paragraphs the first sentence can be treated as the summarization of the rest of the paragraph. \wxb{The above claim is a bit weak. Reviewers may ask for statistical supports.} Due to this common nature of text classification and paragraphs, 
% \isak{too oral and lack motivation - why you have FSP}
% The biggest challenge of applying SSTuning to zero-shot text classification tasks is to find a proper way to construct training samples 
The biggest challenge of applying SSTuning to zero-shot text classification tasks is to design a proper learning objective that can effectively construct large-scale training samples without manual annotations. 
Intuitively, the core of the text classification task can be treated as associating the most suitable label to the text, given all possible options. Motivated by this observation, we propose a new learning objective named first sentence prediction (FSP) for the SSTuning framework to capture such patterns. 
%Given... we propose a new learning objective, First Sentence Prediction (FSP), which enables tuning only with unlabeled data. %out-of-domain \isak{out-of-domain a specific term, avoid this} data.
% Given a paragraph, FSP guides a model to find its first sentence given the rest sentences.%} with the first sentences of other paragraphs as candidates.
% In general, %the first sentence tends to be the most relevant to a paragraph
% the first sentence tends to summarize the main idea of a paragraph, which is similar to the relationship between the label and text in classification tasks.
% % so FSP encourages the model to learn the matching between the first sentence and the rest of a paragraph. 
In general, the first sentence tends to summarize the main idea of a paragraph. Therefore, predicting the first sentence with the rest of the paragraph encourages the model to learn the matching relation between a text and its main idea ("label").
To generate training samples, we use the first sentence in the paragraph as the positive option and the rest as text. The first sentences in other paragraphs are used as negative options. Specifically, if negative options are from the same article as the positive option, they are regarded as hard negatives since the sentences in the same article normally have some similarities, such as describing the same topic. Hard negatives may force the model to learn the semantics of the text instead of simply matching the keywords to complete the task.
% \isak{options, hard negatives should be explained (although briefly). otherwise it is too high-level here}
%
% \isak{In general, the first sentence tends to summarize the main idea of a paragraph. Therefore, predicting the first sentence with the rest of the paragraph can encourage the model to learn the matching relation between a text and its main idea (``label''). In specific, we treat the ... (then describe the details)}

In the inference phase, we convert all possible labels of a sample into options, which can be done in two simple ways: 1) use original label names; 2) convert labels using the templates (like "This text is about [label name]"). Then the text and options are combined to create the final input. The tuned model can thus retrieve the most relevant option as the predicted label of the text.
Since the tuned model has seen a large number of samples and various first sentences as options, which has a higher chance to consist of similar options to the ones at the inference phase,
% it is easier and more flexible to design a proper verbalizer.
its performance is less sensitive to verbalizer design.
In this way, our SSTuning enables efficient deployment of PLM for classifying texts of unseen classes on-the-fly without requiring further tuning with labeled data or unlabeled in-domain data.
% With experiments on 10 popular classification tasks, we demonstrate that our approach outperforms the state-of-the-art baselines.
% \isak{The core of the text classification task can be treated as associating the most suitable label to the text, given all possible labels. Motivated by this observation, we propose a new learning objective named first sentence prediction (FSP) during the SSTuning framework to capture such patterns. Given...}

Our main contributions are:% \isak{the first two are talking about the same thing}:
\begin{itemize}
    % \item We propose a new learning paradigm called Self-Supervised Tuning to solve zero-shot text classification tasks, with a new learning objective called First Sentence Prediction. 
    \item We propose a new learning paradigm called self-supervised tuning (SSTuning) to solve zero-shot text classification tasks. A simple yet effective learning objective named first sentence prediction is designed to bridge the gap between unlabeled data and text classification tasks.
    % to generate tuning and validation data. 
    %\isak{not from ``data'' perspective, but from ``task''}\cq{done}
    %capture the xxx
    % \item We propose a new method to construct datasets from the unlabelled corpus. The datasets will be used for both training and validation.
    % \item We conduct extensive experiments and show that our approach outperforms the baselines, demonstrating that Self-Supervised Tuning is efficient to solve zero-shot classification tasks.
    \item We conduct extensive experiments on 10 zero-shot text classification datasets. The results show that SSTuning outperforms all previous methods on overall accuracy in both topic classification tasks and sentiment analysis tasks. Our analysis further demonstrates that our model is less sensitive to prompt design.
    %The code and tuned models will be publicly available. %(or talking about the average improvements), 
    % \isak{delete last sentence and put it into abstract ending}
\end{itemize}

\section{Proposed Method}
% \isak{
% % 1. Reorganize: 2.3 -> 2.1+2.2 -> 2.4
% 2. highlight the methodology and hide the details (those should be moved to the experimental section
% 3. too many Yang et al
% }

\begin{figure*}[ht]
    \centering
    \includegraphics[width=1\linewidth]{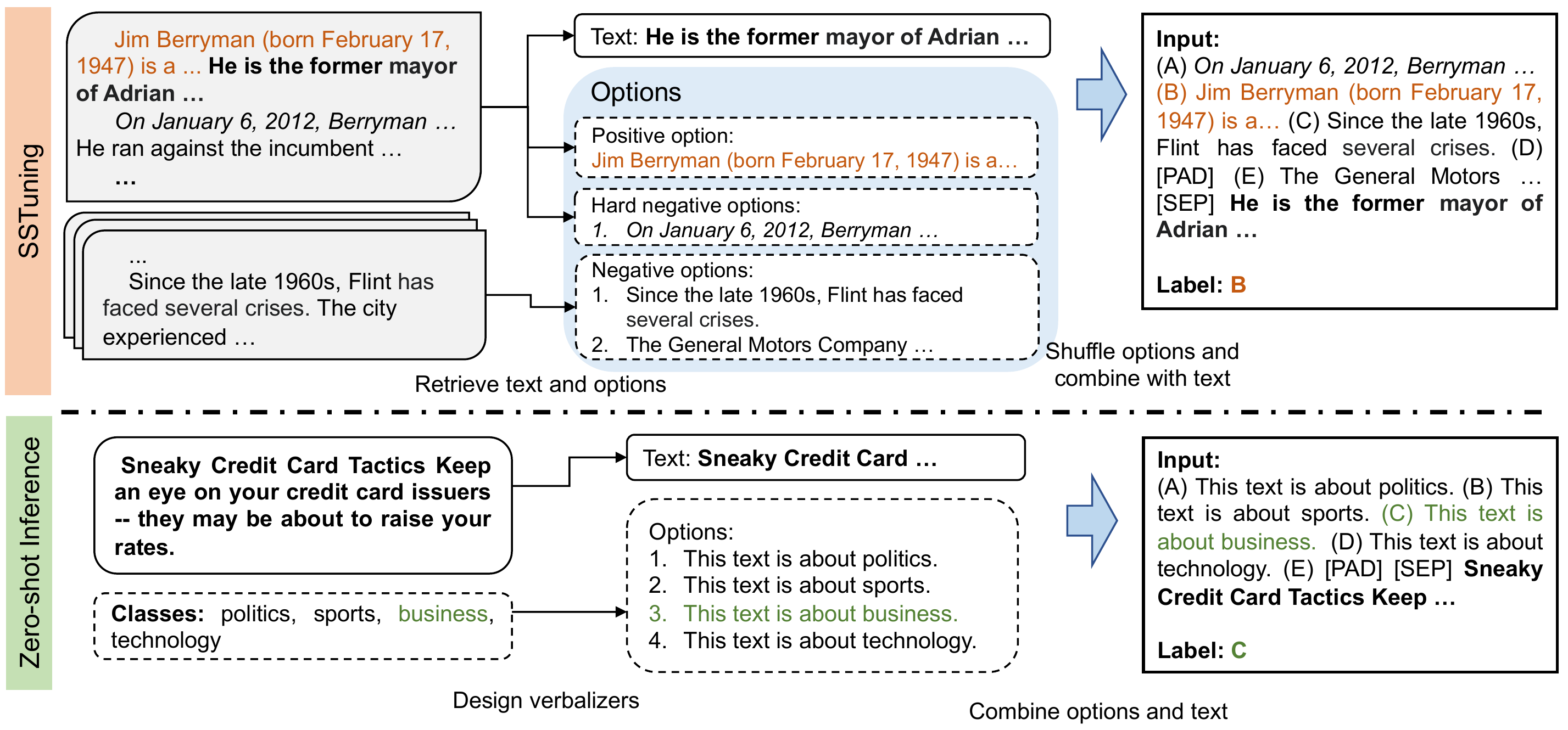}
    \caption{Data construction for SSTuning (top) and zero-shot inference (bottom). The number of labels $N_{\text{model}}$ is set as 5 here. The SSTuning example is from Wikipedia and the inference example is from AG News dataset. }% The learning objective during training is to predict the index of the corresponding first sentence \underline{underlined} for the rest of a paragraph \textbf{in bold}. The rest of the options are either first sentences from different paragraphs or pad tokens ([PAD]). First sentences from the same article but another paragraph (\textit{in italic}) is treated as hard negatives. During inference, the class labels are converted to more meaningful verbalizers and the input is transformed into the same format with training. The target will be the index of the correct label. The three examples are from Wikipedia corpus, AG News datasets, respectively. \isak{many irrelevant/less important details, making the focus unclear: shorten the caption and make the figure clearer}}
    \label{fig:data_gen}
\end{figure*}

% In this section, we discuss the proposed framework SSTuning, including the dataset generation based on first sentence prediction (FSP), the tuning phase, and the zero-shot inference phase in detail. 
In this section, we discuss our proposed framework, SSTuning, and provide details for our dataset preparation process using the idea of first sentence prediction (FSP), the tuning phase, and the zero-shot inference phase.
% \isak{
% 1. better from ``task construction'' perspective, not ``data construction''
% 2. motivation is missing - why FSP
% 3. use the figure to demonstrate}

% \subsection{The UniClassify Framework}
% The proposed method is shown in Figure \ref{fig:uniclassify}. Compared with traditional pre-training fine-tuning approaches, we have made the following changes: i) we reformulate the input to include the label information; ii) instead of training with labeled in-domain data, we train with the unlabeled out-of-domain corpus; iii) we change the learning objective. \isak{already in the model section, no need to compare, just describe}

% \subsection{SSTuning Dataset Generation \isak{FSP}}
\subsection{First Sentence Prediction}
% First sentence in a paragraph tends to be more relevant to the rest of the paragraph than first sentences in other paragraphs. This is similar to the text classification task that aims to select the most relevant label for the text. \guizhen{hmm i dont rly see the links between the first 2 sentences.} 
Text classification can be regarded as selecting the most relevant label for the text, given all possible labels. 
Based on such observation, we propose the FSP task to create datasets for our SSTuning by mimicking the same structure.

% For a paragraph, we use the first sentences as its positive option and the first sentences of other paragraphs as negatives. 

% \wxb{Try to avoid passive sentences as much as possible.} --> OK
We design the FSP task by considering both the nature of the unlabeled corpus and the input/output format of classification tasks. In this subsection, we describe in detail how to construct the tuning and validation sets from the unlabeled corpus. Figure \ref{fig:data_gen} shows the core procedures for our dataset generation.

% % \wxb{Use the paragraph command for small titles.}
% \paragraph{Selection of unlabeled data.}
% The unlabeled data should have the feature that the first sentence in a paragraph has a strong correlation with the rest of the paragraph. For example, Wikipedia articles normally talk about one specific topic, making it a possible candidate. 
% \wxb{This paragraph seems useless.} \isak{agree}

\paragraph{Data filtering.}
% \guizhen{should we remove the full stop and capitalize all first letters?} \cq{I think no. Please check \citep{emnlp22/mining-based}}
We first filter data to select appropriate paragraphs for tuning (more details are shown in \ref{sub_sec:data_filter}). Removing meaningless sentences ensures data quality, which helps improve the performance of the model.
% Not all paragraphs in a selected corpus are eligible to create samples. 

% \wxb{It's too vague. Explain the detail here or mention that please see the detail in the appendix.} \isak{agree, some of them are too detailed (and obvious)}
% \isak{remove the above two parts (some can be moved to exp details) and replace it as motivation}

\paragraph{First sentence as the positive option.}
% \wxb{Consider using positive option since the following uses negatives.}
% \guizhen{"Let's" is informal.}
% Let's assume we have a collection of articles %[$A_1, A_2, ... A_N$] 
% after data filtering, and article $A_n$ has $M$ paragraphs [$P_1^n,P_2^n,... P_M^n$].
% % \wxb{$A_1$ isn't used in the following. Consider removing it.}
% Consider a paragraph $P_m^n$ that includes $K$ sentence [$S_1^{n,m},S_2^{n,m},...,S_K^{n,m}$].
% The positive option $O_c^{n,m}$ and the text $x^{n,m}$ are: \guizhen{
We consider an article $A_n$ that contains $M$ paragraphs, i.e., $A_n = [P_1^n,P_2^n,... P_M^n]$, and suppose paragraph $P_m^n$ has $K$ sentences $[S_1^{n,m},S_2^{n,m},...,S_K^{n,m}]$, the positive option $O_c^{n,m}$ and the text $x^{n,m}$ are:%}
\begin{align}
    O_c^{n,m} &= S_1^{n,m} \\
    x^{n,m} &= [S_2^{n,m},...,S_K^{n,m}]
\end{align}
As shown in Figure \ref{fig:data_gen}, we can retrieve the first sentence \textit{"Jim Berryman (born February 17, 1947) is a ... "} as the positive option and the rest of the paragraph \textit{"He is the former mayor of Adrian ..."} as the text for the first paragraph in the article.% \textit{"Jim Berryman"}.

\paragraph{Negative sampling.} After getting the positive option, we randomly sample $J$ "first sentences" from other paragraphs [$S_1^{n_1,m_1},S_1^{n_2,m_2},... S_1^{n_J,m_J}$] as negative options, where $J$ is a random number that satisfies $1 \leq J \leq N_{\text{maxLabel}}-1$. We let $N_{\text{maxLabel}}$ denote the maximum number of labels that are first sentences, which is pre-defined to ensure the total number of tokens for options is not too long. It is less or equal to $N_{\text{model}}$, where $N_{\text{model}}$ is the number of labels for the model output layer. 
% There are two reasons for having a random number of negative options. 1) During inference, the number of classes is not fixed, which may vary from $2$ to $N_{\text{model}}$. %\wxb{Also use \textbackslash text for these subscripts.} 
% 2) It is easier to train with a small number of negative options and harder with a large number. 
Having a random number of negative options bridges the gap between tuning and zero-shot inference since the number of classes for evaluation datasets may vary from $2$ to $N_{\text{model}}$.
% Our preliminary experiment shows that a mixing number of negatives will get both stable training and good performance (See sectionxxx).

\paragraph{Hard negatives.} During negative sampling, if the negative options and the positive option are from the same article, we call the options hard negatives. Inspired by the successful application of hard negatives in \citet{emnlp21/simcse}, we purposely add more hard negatives to enhance the model performance. 
% In some articles, we observe that in the same paragraph, common words are likely to appear in the first sentence and the rest of the paragraph at the same time. As shown in Figure \ref{fig:data_gen}, \textit{"Berryman"} can be a shortcut to select the corresponding first sentence for the text. However, if we add the hard negative \textit{"On January 6, 2012, Berryman ..."}, the model needs to understand the true semantics to select the positive option.
Sometimes, when we read articles, we notice that the same words appear in the first sentence and the rest of the paragraph. As shown in Figure \ref{fig:data_gen}, we can use the word \textit{"Berryman"} to quickly find the corresponding first sentence for the text. However, if we add the hard negative \textit{"On January 6, 2012, Berryman ..."}, the model has to understand the true semantics to choose the positive option.

\paragraph{Option padding.}
We pad the options with the special "\texttt{[PAD]}" token to make the input format consistent between the tuning phase and the inference phase.
% \wxb{It seems we usually use \textbackslash texttt for special tokens like PAD, CLS.}
% This is because we need to make the input format consistent between the tuning phase and the inference phase.
% To make the input format consistent between the tuning phase and the inference phase, we pad the options with "[PAD]" token.
Specifically, if the total number of options after negative sampling is $(J+1) < N_{\text{model}}$, we will add $(N_{\text{model}} - J -1)$ \texttt{[PAD]} options. Thus the final list of options is: 
\begin{equation}
\begin{aligned}
    O^{n,m} = [S_1^{n,m},S_1^{n_1,m_1},S_1^{n_2,m_2},...S_1^{n_J,m_J},\\
    O_{\texttt{PAD}}^1,O_{\texttt{PAD}}^2,...O_{\texttt{PAD}}^{N_{\text{model}} - J -1}]
\end{aligned}
\end{equation}

\paragraph{Generating final text and label.} 
We shuffle the option list because the position of a positive option is random in the evaluation datasets.
% Since the correct option may appear anywhere in the test dataset, we shuffle the option list to mimic the same setting.
% Assume the option list after shuffling is:
After shuffling, we assume the option list is:
\begin{equation}
    O^{n,m}_{\text{shuffle}} = [O_0,O_1,...O_{N_{\text{model}}-1}],
\end{equation}
where the positive option $O_c^{n,m} = O_{j}$.
Then the label for this sample is: 
\begin{equation}
    L^{n,m} = j.
\end{equation}
The final input text is the concatenation of the above components: 
\begin{equation}
    x_{inp}^{n,m} = [\text{CLS}]\{(T_i)\ O_{i}\}_{i=0}^{N_{\text{model}}-1}[\text{SEP}]x^{n,m}[\text{SEP}]
\end{equation}
where $T_i$ is the $i$-th item from the index indicator list $T$ (e.g. $[A,B,C...]$), \texttt{[CLS]} is the classification token, and \texttt{[SEP]} is the seperator token used by \citet{naacl19/bert}.
% \wxb{This equation lacks concat operations.}

Thus the final text-label pair ($x_{inp}^{n,m}$, $ L^{n,m}$) is the generated sample. We can repeat this process to generate a large number of samples as the tuning set. The validation set can also be generated in the same way.
Note that if we select a corpus that only contains paragraphs instead of articles, we can treat each paragraph as an article, and no hard negatives are generated. 
% \wxb{Use present tense as much as possible.}

\subsection{Tuning Phase}
% \guizhen{\st{During the tuning phase, we tune the model with the generated datasets, just like normal fine-tuning for BERT. }}
\subsubsection{Network Architecture}
We employ BERT-like pre-trained masked language models (PMLM) as the backbone, such as RoBERTa \citep{corr/roberta} and ALBERT \citep{iclr20/albert}. Following \citet{naacl19/bert}, we add an output layer for classification. Such models have both bidirectional encoding capabilities and simplicity. Generative models are not necessary since we only need to predict the index of the correct option.  We do not make any changes to the backbone so that the method can be easily adapted to different backbones. In order to cover all test datasets, we config the number of labels for the output layer as the maximum number of classes for all test datasets, denoted by $N_{\text{model}}$. 

\subsubsection{Learning Objective}
Traditional text classification with PMLMs like BERT maps each classification layer output to a class. Such a design requires a dedicated output layer for each dataset as they have different classes. Instead, our learning object for FSP with the same network is to predict the index of the positive option. In this way, we can use the output layer for both tuning and inference and for various kinds of datasets.

% \wxb{Restrain from using 'just'. This may weaken the novelty of this paper.}
As shown in Figure \ref{fig:data_gen}, we concatenate the labels and the text as input. %\guizhen{We use cross-entropy loss because ....} 
The outputs are the indices (0, 1, 2..., which correspond to A, B, C), which are the same as traditional classification datasets. We use a cross-entropy loss for tuning the model.%} \guizhen{avoid using too many 'just'} 
% When inference with the same text but different label orders, the target may be different. For the following two examples: 
% \begin{enumerate}
%     \item "(A) Positive. (B) Negative. [SEP] What a wonderful movie."
%     \item "(A) Negative. (B) Positive. [SEP] What a wonderful movie."
% \end{enumerate}

\subsection{Zero-Shot Inference Phase} %\guizhen{zero-shot phase sounds abit weird}
During the zero-shot inference phase, we can infer directly by converting the input of the sample to the same format as that in the tuning phase. 

\subsubsection{Input Formulation} \label{sec:input}
% Instead of using the original text as the input to the model, we include the label information in the input. 
As shown in Figure \ref{fig:data_gen}, the zero-shot inputs are formulated similarly as the tuning phase, except 1) instead of using first sentences as options, we convert the class names to options. Actually, we can simply use the original labels or some simple templates like “This text is about [label name].” for the conversion, thus little to no effort is needed. 2) No shuffling is needed. Since the converted input and output during SSTuning and zero-shot phases are the same, no further adjustment of the model is required.

% Given an input example $x$ and the label space $L = L_1, L_2 ... L_{|L|}$, we first map each label to a string $v: L \rightarrow V$, where $|V|$ = $|L|$. To have a unified format for input, we fix the total number of classes to $N$, where $N$ should be larger or equal to $|V|$. Then we add $(N -|V|)$ pad labels. We will shuffle the label list during training time but not during inference time. The final label list will be $V_F$. Assume we choose the tag list $T$ (e.g. $[A,B,C...]$), then the final input $x_{inp}$ is formulated as: 
% \begin{equation}
%     x_{inp} = [\text{CLS}]\{(T_i)\ V_{Fi}\}_{i=1}^{N}[\text{SEP}]x[\text{SEP}]
% \end{equation}

\subsubsection{Constrained Prediction}
Since the dimension of the output logits ($N_{model}$) may be different from the number of classes in a dataset ($N_L$), the predictions may be out of range (e.g. the model may output 3 for a dataset with 2 classes). To solve this issue, we simply make predictions based on the first $N_L$ logits: 
\begin{equation}
    P = \text{argmax}(\text{logits}[0:N_L])
\end{equation}
where $P$ is the index for the positive option.
% \wxb{Exactly explain what is $P$ here.}

\section{Experiment Setup}
% \isak{reorganize this section by their importance: 3.2 -> 3.3 -> 3.1 -> 3.4 Experiment Details (this part)}
% \isak{be careful about the format, if use $1$ for numbers, then all numbers better in this form. otherwise just all numbers use normal form: 1}

\subsection{SSTuning Datasets}%\subsection{Training and validation datasets} \isak{SSTuning Datasets (at least avoid ``training'')}
We choose English Wikipedia %\citep{wikidump} 
and Amazon review dataset (2018) \citep{emnlp19amazon_review} for SSTuning. %The reason for choosing the two datasets are: \isak{can delete the above sentence, directly talking about the advantage of using them} 1) 
The Wikipedia corpus has more than 6.2M articles\footnote{\url{https://en.wikipedia.org/wiki/Wikipedia:Size_of_Wikipedia}} by the end of 2021, while Amazon Review Data has around 233.1M reviews\footnote{\url{https://nijianmo.github.io/amazon/}}. 
%Wikipedia use formal text and Amazon review use informal text, covering a broader range of scenarios. \isak{
Wikipedia articles typically use formal expressions and Amazon reviews contain informal user-written texts, together covering different genres of text.

For English Wikipedia, we collect articles up to March 1st, 2022.  To balance the dataset, we select up to 5 paragraphs in each article. The generated dataset has 13.5M samples. %After spliting, we have 13.4 million samples as the training set and 0.1 million samples as the validation set.
% \isak{the above two paragraphs are "experiment details", not "methodology"}
For the Amazon review dataset, we only use the review text 
% \isak{what is ``main review'', readers do not know amazon contains which part} 
to create our SSTuning dataset, ignoring other information such as summary and vote. The Amazon review dataset has 29 categories. To keep the model from being dominated by a certain category, we select up to 500k samples from each category. In the end, we collected 11.9M samples. %After spliting, we have 11.8 million samples for training set and 0.1 million samples for validation set. %We tried using the reviews from the same categories as hard negatives, but there is no meaningful improvement for the performance. Thus we do not apply hard negatives for the Amazon review dataset. \isak{last sentence can be moved into experiment detail section (since this has nothing to do with ``data'')}

To have a balanced dataset, we sample 2.56M from the Wikipedia dataset and 2.56M from the Amazon review dataset, forming a total of 5.12M samples as the tuning dataset. In addition, we sampled 32k from each of the two datasets, forming a validation set consisting of 64k samples.
% \isak{describe the final used dataset size [important]}
% \cq{The following paragraphs need to be filtered out: 1) Paragraphs with a single sentence, otherwise they cannot be split into option and text. 2) Paragraphs with meaningless first sentences, since some sentences may only contain meaningless characters. However, the corpus may still be noisy after filtering.}

\subsection{Evaluation Datasets}
% We evaluate the models on 4 topic classification (TC) tasks and 6 sentiment analysis (SA) tasks. Topic classification tasks include Yahoo Topics (yah) \citep{nips15/amazon_yelp_SA}, AG News (agn) \citep{nips15/amazon_yelp_SA}, DBPedia (dbp) \citep{nips15/amazon_yelp_SA} and 20newsgroup (20n) \citep{icml95/20news}. Sentiment analysis tasks include SST-2 (sst2) \citep{emnlp13/sst2_SA}, IMDb (imd) \citep{acl11/imdb}, Yelp (ylp) \citep{nips15/amazon_yelp_SA}, MR (mr) \citep{acl05/MR_SA} and Amazon (amz) \citep{nips15/amazon_yelp_SA}, which are binary classification tasks, and SST-5 (sst5) \citep{emnlp13/sst2_SA}, which is a fine-grained classification task. Detailed data statistics for each testing dataset are presented in Table \ref{tab:stats} in Appendix \ref{sec:appendix}.

We evaluate the models on 4 topic classification (TC) tasks, including Yahoo Topics (yah) \citep{nips15/amazon_yelp_SA}, AG News (agn) \citep{nips15/amazon_yelp_SA}, DBPedia (dbp) \citep{nips15/amazon_yelp_SA} and 20newsgroup (20n) \citep{icml95/20news}, and 6 sentiment analysis (SA) tasks, including SST-2 (sst2) \citep{emnlp13/sst2_SA}, IMDb (imd) \citep{acl11/imdb}, Yelp (ylp) \citep{nips15/amazon_yelp_SA}, MR (mr) \citep{acl05/MR_SA} and Amazon (amz) \citep{nips15/amazon_yelp_SA}, which are binary classification tasks, and SST-5 (sst5) \citep{emnlp13/sst2_SA}, a fine-grained 5-class SA task. Detailed data statistics for each testing dataset are presented in Table \ref{tab:stats} in Appendix \ref{sec:appendix}.

Following the baselines \citep{emnlp22/uniMC,emnlp22/mining-based,emnlp22/self-training}, we report the accuracy on the test set when available, falling back to the original validation set for SST-2.

\subsection{Baselines}
% \isak{no need to introduce the baselines too detailed, shorten this part}
We choose the following baselines for comparison after considering their relevancy, impact, checkpoint availability, and model sizes: %\wxb{Consider using enumerate/itemize for listing baseline models.}

\begin{itemize}
    \item Textual entailment (TE) \citep{emnlp19/TE}: Following \citet{emnlp22/self-training}, we download the off-the-shelf models trained on MNLI and use the default hypothesis template \textit{"This example is []."} for evaluation.
    % and rely on the Hugging Face Transformers library\footnote{\url{https://huggingface.co/zero-shot/}} for inference, 
    % We choose \textit{roberta-large-mnli} and \textit{bart-large-mnli} to represent this baseline.
    \item \text{TE-Wiki \citep{naacl22/TE_TowardsOT}:} This model is also trained with entailment methods but with a dataset constructed from Wikipedia. %The model is specially trained for topic classification. We also download the off-the-shelf model from Hugging Face\footnote{\url{https://huggingface.co/CogComp/ZeroShotWiki}}. 
    \item Prompting-based method \citep{NAACL21/iPET}: We compare with the results using multiple verbalizers reported in \citep{emnlp22/mining-based}.
    \item \text{Mining-based \citep{emnlp22/mining-based}:} The method has three steps, which are \textit{mine}, \textit{filter} and \textit{fine-tune}. 
    % This can be regarded as a weakly-supervised method. 
    We compare with the results reported.
    \item \text{UniMC \citep{emnlp22/uniMC}:} We download the released checkpoint 
    % from Hugging Face\footnote{\url{https://huggingface.co/IDEA-CCNL/Erlangshen-UniMC-Albert-235M-English}} to test the datasets. We
    and test the model without question prompts since the reported results on text classification tasks are better on average. %We also compare with the results reported in the paper, as the reported result is based on $5$ seeds and our reproduced result is based on only $1$ seed.
\end{itemize}

We followed the setups and verbalizers of the original works as much as possible. If the original work does not have verbalizers for a dataset, we will use the same or comparable verbalizers as ours, as shown in Table \ref{tab:verbalizer}.

\subsection{Implementation Details}
To test the performance of the proposed method on different model sizes and architectures, we tune three versions of models, which are based on $\text{RoBERTa}_{\text{base}}$, $\text{RoBERTa}_{\text{large}}$ \citep{corr/roberta}, and $\text{ALBERT}_{\text{xxlarge}}$ (V2) \citep{iclr20/albert}, denoted as SSTuning-base, SSTuning-large, SSTuning-ALBERT, respectively. We set the maximum token length as 512 and only run one epoch. We repeat all the experiments 5 times with different seeds by default. The experiments on SSTuning-base and SSTuning-large are run on 8 NVIDIA V100 GPUs and the experiments on SSTuning-ALBERT are run on 4 NVIDIA A100 GPUs. 

The hyperparameters for fine-tuning and SSTuning are shown in Table \ref{tab:hyperparameters}. We set the batch size based on the constraint of the hardware and do a simple hyperparameter search for the learning rate. %\isak{discuss how the hyperparameter are selected here, details can be put into the appendix}
We do not add hard negatives for the Amazon review dataset since the reviews are not in the format of articles. We also tried to use the negative options from the same product category as hard negatives  but did not find any meaningful improvement. 
%\isak{can mention empirical evidence}
% We set the maximum number of classes $N_\text{model}$ for SSTuning output layers as 20, since 20newsgroup has the maximum number of classes 20 amount all the evaluation datasets. We set $N_{\text{maxLabel}}$ as 10 after simple experiment.
We set $N_\text{model}$ as 20 and $N_{\text{maxLabel}}$ as 10 after simple experiment.

\section{Results and Analysis}
% In this section, we will present the performances of our models on the 10 datasets and the ablation studies. \isak{useless, can remove (we normally use this kind of sentence when the section is very long containing different kinds of topics, so here it is unnecessary)}
% \isak{haven't revised the expressions, adjust the organization first}

\subsection{Main Results} \label{sec:main_results}
% \isak{MARK: Note we often first fully show the advantage of our method, then discuss the weakness. imbalance here}
\begin{table*}[!ht]
    \centering
    % \setlength\tabcolsep{1pt}
    % \small
    \resizebox{\textwidth}{!}{%
    \begin{tabular}{llcccccccccccccc}
    \toprule
        & \multirow{2}{*}{\textbf{Backbone}} & \multirow{2}{*}{\textbf{Labeled}} & \multicolumn{5}{c}{\textbf{Topic Classification}} &\multicolumn{7}{c}{\textbf{Sentiment Analysis}} & \multirow{2}{*}{\textbf{Avg}} \\
        % \cline{4-7} \cline{9-14}
        \cmidrule{4-7} \cmidrule{9-14}
         &  &  & \textbf{yah} & \textbf{agn} & \textbf{dbp} & \textbf{20n} & & \textbf{sst2} & \textbf{imd} & \textbf{ylp} & \textbf{mr} & \textbf{amz} & \textbf{sst5} & & \\ 
         \midrule
        Fine-tuning\textsuperscript{\ding{118}} &RoBERTa\textsubscript{\text{large}} & -   & 77.1 & 95.5 & 99.2 & 75.3 & & 95.9 & 96.4 & 98.3 & 91.3 & 97.2 & 59.9 & & 88.6 \\ \midrule
        % TE-Wiki* &110M  & \cmark      & 57.3 & 79.6 & 90.2 & - & & - & - & - & - & - & - & & - \\
        TE-Wiki &BERT\textsubscript{\text{base}} & \cmark      & 56.5 & 79.4 & 90.4 & 53.9 && 57.3 & 62.0 & 58.5 & 56.2 & 55.8 & 24.5 && 59.5 \\ 
        TE-MNLI &RoBERTa\textsubscript{\text{large}} &\cmark  & 28.6 & 77.6 & 60.4 & 40.2 && 89.6 & 90.2 & 92.8 & 82.8 & 92.0 & \textbf{48.8} && 70.3 \\ 
        TE-MNLI &BART\textsubscript{\text{large}} &\cmark     & 48.2 & 74.8 & 57.1 & 35.4 && 89.0 & 91.1 & 93.1 & 81.4 & 91.9 & 47.7 && 71.0 \\ 
        Prompting* &RoBERTa\textsubscript{\text{base}} &- & 34.1 & 54.6 & 51.1 & -& & 81.9 & 81.8 & 83.1 & 78.3 & 83.5  & - & & - \\
        Mining-based* &RoBERTa\textsubscript{\text{base}} &\xmark  & 56.1 & 79.2 & 80.4 & - & & 85.6 & 86.7 & 92.0 & 80.5 & 92.0 & - & & - \\ 
        % FLAN &137B & \cmark         & - &- & - & - & 92.6* & 94.1* & 97.8* & - & - & - & - \\ 
        UniMC* &ALBERT\textsubscript{\text{xxlarge}} & \cmark        & - & 81.3 & 88.9 & - & & \textbf{91.6} & \textbf{94.8} & - & - & - & - & & - \\ 
        % UniMC &ALBERT_{\text{xxlarge-V2}} & \cmark        & 63.3 & 85.2 & 62.3 & 51.1 & & 90.9 & 94.5 & 96.1 & 89.0 & 94.8 & 42.9 & & 77.0 \\ 
        UniMC (Rerun) &ALBERT\textsubscript{\text{xxlarge}} & \cmark        & 59.0 & 84.3 & 89.2 & 43.7 && 90.1 & 93.6 & 94.3 & 87.3 & 93 & 45.6 && 78.0  \\ 
        % Selft-training &&\xmark & 62 & 81.4 & 94.5 & 65.8 & - & 92.3 & - & - & 95 & - & - \\ 
        \midrule
        SSTuning-base   &  RoBERTa\textsubscript{\text{base}} &\xmark & 59.1 & 79.9 & 82.7 & 47.2 && 86.4 & 88.2 & 92.9 & 83.8 & 94.0 & 45.0 && 75.9 \\ 
                            % &&& ($\pm0.7$) & ($\pm2.5$)$ & ($\pm1.9$)$ & ($\pm1.4$)$ && ($\pm0.7$)$ & ($\pm1.3$)$ & ($\pm0.7$)$ & ($\pm1.1$)$ & ($\pm0.3$)$ & ($\pm0.6$)$ && ($\pm0.3$)$ \\ 
        SSTuning-large  &  RoBERTa\textsubscript{\text{large}} &\xmark & 62.4 & 83.7 & 85.6 & 56.7 && 90.1 & 93.0 & 95.2 & 87.4 & 95.2 & 46.9 && 79.6 \\ 
                            % &&& 0.6 & 0.6 & 1.1 & 1.6 && 0.5 & 0.8 & 0.6 & 0.8 & 0.3 & 2.5 && 0.3 \\
        SSTuning-ALBERT & ALBERT\textsubscript{\text{xxlarge}} &\xmark &\textbf{63.5} & \textbf{85.5} & \textbf{92.4} & \textbf{62.0} && 90.8 & 93.4 & \textbf{95.8} & \textbf{89.5} & \textbf{95.6} & 45.2 && \textbf{81.4} \\ 
                            % &&& 0.7 & 1.0 & 1.4 & 0.7 && 0.2 & 0.6 & 0.5 & 0.5 & 0.3 & 3.5 && 0.4 \\
    \bottomrule
    \end{tabular}
    }
    \caption{Main results for 4 topic classification tasks and 6 sentiment analysis tasks. \ding{118}: the original training sets (see dataset sizes in Table \ref{tab:stats}) are used to provide results under supervised settings, served as upper bound, otherwise zero-shot results are reported. *: results are taken from corresponding papers. "Labeled" indicates whether the model uses labeled (\cmark) or unlabeled (\xmark) data. "Avg" is the arithmetic mean accuracy of all the datasets. For SSTuning models, we report the mean accuracy of 5 runs using different seeds. The best results for each dataset are in bold. %\isak{add a column for the used intermediate dataset (same as excel) would be great, and even more important than number of parameters}
    }
    \label{tab:main_results}
\end{table*}

The main results are shown in Table \ref{tab:main_results}. We have the following observations: 
1) Our method SSTuning-ALBERT achieves new state-of-the-art results on 7 out of 10 datasets, and
significantly reduces the gap between fine-tuning and zero-shot methods compared to UniMC (from 10.6 to 7.2)
% even approaches the performance with the supervised setting (i.e., results of fine-tuning)
, showing the superiority of our proposed method.
% 2) Compared to UniMC with SSTuning-ALBERT using the same backbone, we still outperform it by 3.4\% on average. 
2) With the same backbone, SSTuning-ALBERT outperforms UniMC by 3.4\% on average.
Note that different from UniMC, we do not utilize any labeled data to conduct meta-tuning, but purely rely on auto-constructed data for self-supervised tuning, which not only has a much large scale of data but also has more abundant options (first sentences).
3) Comparing methods based on RoBERTa\textsubscript{\text{base}}, RoBERTa\textsubscript{\text{large}} and BART\textsubscript{\text{large}}, our SSTuning-large and SSTuning-base are the two best-performing models on average. We also observe that SSTuning-large outperforms UniMC, despite the latter possessing a stronger backbone.
4) Our models do not perform very well on SST-5, which is a fine-grained sentiment analysis task. Maybe we can generate more fine-grained options from the unlabeled corpus to improve performance on such tasks. We leave it as a future work.
%talk about the weakness, but from the ``extension'' perspective

\begin{table}[]
    \centering
    \small
    % \resizebox{\linewidth}{!}{
    \begin{tabular}{lccc}
    \toprule
         & \textbf{TC} & \textbf{SA} & \textbf{All} \\
        \midrule
        Amazon  &  63.4 & 81.4 & 74.2 \\
        Wikipedia &  63.4 & 77.9 & 72.1 \\ 
        Amazon + Wikipedia & \textbf{67.2} & \textbf{81.7} & \textbf{75.9} \\
        % Amazon  &  63.7 & 80.2 & 73.6 \\
        % Wikipedia &  63.4 & 65.1 & 64.4 \\ 
        % Amazon + Wikipedia & \textbf{67.1} & \textbf{80.9} & \textbf{75.4} \\
    \bottomrule
    \end{tabular}
    % }
    \caption{Zero-shot results with different tuning datasets. The best result is in \textbf{Bold}. }
    \label{tab:impact_dataset}
\end{table}

% \subsubsection{Further training with target domain labeled data}
% train with unlabeled target training data
% % \subsection{Few-shot setting}

\subsection{Ablation Study}
% \isak{use \ paragraph{} for each subsection, and shorten the text content in each part.}
%\subsubsection{How does the training set impact the performance?}
% \paragraph{Ablation on Tuning Datasets}
\subsubsection{Ablation on Tuning Datasets}
% In the main experiment, we trained the models with both the Amazon review dataset and the Wikipedia corpus. To evaluate whether the two datasets are necessary, we also trained with them separately. The total number of samples for each case is 5.12M for a fair comparison. The results are shown in Table \ref{tab:impact_dataset}. We calculate the mean accuracy for all the topic classification tasks, all the sentiment analysis tasks, and all the tasks, respectively. It shows that tuning with both datasets get the best results for all the 3 cases. 
% It is interesting to see that tuning with Amazon Product Review performs the same as tuning with Wikipedia corpus on topic classification tasks. This is unexpected since Wikipedia is more related to topic classification tasks intuitively. This may be because the models have already been pre-trained with Wikipedia, thus further tuning with it does not bring more advantages. 
We utilize both the Amazon review dataset and English Wikipedia during the tuning stage. To evaluate their effectiveness, we conduct ablation studies to create two model variants that are only trained on one dataset. We set the number of samples for each case to 5.12M for a fair comparison. As shown in Table \ref{tab:impact_dataset}, both datasets contribute to the final performance, thus discarding any one leads to a performance drop. It is interesting that tuning with Amazon review data performs the same as tuning with Wikipedia on topic classification tasks. This is unexpected since Wikipedia is more related to topic classification tasks intuitively. We anticipate the reason is that the backbone models have already been pre-trained with Wikipedia, thus further tuning with it does not bring significant advantages.

% \subsection{Analysis}
%\subsubsection{Can we predict other sentences in a paragraph as labels?}
% \paragraph{Alternative Tuning Objectives}
\subsubsection{Alternative Tuning Objectives}
% Like FSP, we can also use other sentences in a paragraph as the training signal. We consider the following settings: 1) last-sentence prediction (LSP), which treats the last sentence as the label for the rest of the paragraph; 2) next-sentence selection (NSS), which treats the first sentence in a consecutive sentence pair as text and the next as the label; 3) Random sentence prediction (RSP), which randomly pick a sentence in a paragraph as a label and treat the rest as text. The comparison between the four settings is shown in Table \ref{tab:LSP}.
\begin{table}[]
    \centering
    \small
    % \resizebox{\linewidth}{!}{
    \begin{tabular}{lccc}
    \toprule
         & \textbf{TC} & \textbf{SA} & \textbf{All} \\
        \midrule
        First sentence prediction  & \textbf{67.2} & 81.7 & \textbf{75.9} \\
        Last sentence prediction &  59.8	& \textbf{82.2}	& 73.3  \\ 
        Next sentence selection &  54.8	& 81.9	& 71.1 \\
        Random sentence prediction & 56.8	& 80.8	& 71.2 \\
    \bottomrule
    \end{tabular}
    % }
    \caption{Zero-shot results with different tuning objectives. 
    % The mean accuracies for 4 topic classification tasks, for 6 sentiment analysis datasets and for all datasets over 5 seeds are reported. 
    The best results are in \textbf{Bold}.}% \isak{can use the full name of FSP etc since the current table is small, also the reader may not be very familiar with those abbrs.}}
    \label{tab:LSP}
\end{table}

% The result shows that FSP performs the best, especially for topic classification tasks. The second best setting is LSP, which is expected since the last sentences in certain paragraphs are also important. Unlike topic classification tasks, the four settings perform similarly on sentiment analysis tasks. The possible reason is that each sentence in a paragraph shares the same sentiment.

We have proposed first sentence prediction (FSP) as the tuning objective to equip the model learning to associate the label and text in the inference stage. We consider some alternative objectives here for comparison: 1) last sentence prediction (LSP), which treats the last sentence as the positive option for the rest of the paragraph; 2) next sentence selection (NSS)\footnote{Note that we use NSS here to distinguish from NSP (next sentence prediction) used by \citet{naacl19/bert}.}, which treats the first sentence in a consecutive sentence pair as text and the next as the positive option; 3) random sentence prediction (RSP), which randomly pick a sentence in a paragraph as the positive option and treat the rest as text. The comparison between the four settings is shown in Table \ref{tab:LSP}. We find that FSP performs the best, especially for topic classification tasks. Among the alternatives, utilizing LSP as the tuning objective leads to the best performance, which is expected since the last sentence in a paragraph usually also contains the central idea, sharing a similar function as the first sentence. Unlike topic classification tasks, the four settings perform similarly on sentiment analysis tasks. The possible reason is that each sentence in a paragraph shares the same sentiment.

% \isak{add a new section called analysis for the following experiments}
\subsection{Analysis}
% \subsubsection{How does the number of hard negatives impact? \isak{Impact of Hard Negative Samples}}

\begin{table*}[th]
    \centering
    % \small
    \resizebox{0.9\textwidth}{!}{
    \begin{tabular}{llccccc}
    \toprule
         \multirow{2}{*}{\textbf{Verbalizer for "negative"}} & \multirow{2}{*}{\textbf{Verbalizer for "positive"}} & \multicolumn{3}{c}{\textbf{UniMC(w/o Qn)}} & \multicolumn{2}{c}{\textbf{SSTuning-ALBERT}}\\
         \cmidrule{3-4} \cmidrule{6-7}
         & & \textbf{SST-2} & \textbf{IMDb} & & \textbf{SST-2} & \textbf{IMDb} \\
         \midrule
         Bad. & Good. & 87.0 & 91.9 & & 90.7 & 93.9  \\
         Terrible. & Great. & 88.5 & 91.7 & & 91.4 & 94.3 \\
         Negative. & Positive. & 86.0 & 90.3 & & 92.2 & 92.6 \\
         Negative! & Positive! & 88.9 & 90.2 & & 92.1 & 92.4 \\
         Terrible! & Awesome! & 88.4 & 91.1 & & 90.9 & 94.0 \\
         Bad, terrible and negative. & Good, great, and positive. & 80.7 & 87.5 & & 87.3 & 90.8 \\
         I don't like the movie! & I like the movie! & 91.5 & 92.9 & & 89.8 & 90.3 \\
         Terrible! & I like the movie! It is wonderful! & 66.4 & 75.1 & & 86.8 & 92.1 \\
         It's terrible. & It's great. & 91.6 & 93.0 & & 90.6 & 94.1 \\
         It's negative. & It's positive. & 85.6 & 89.9 & & 89.2 & 91.3 \\
         \midrule
         % \midrule
         & Average & 85.5 & 89.4 & & \textbf{90.1} & \textbf{92.6} \\
         & Standard Deviation & 7.4 & 5.3 & & \textbf{1.9} & \textbf{1.5} \\
    \bottomrule
    \end{tabular}}
    \caption{Comparison of zero-shot results for 2 sentiment analysis tasks with different verbalizers. The best average results are in \textbf{bold}.}
    \label{tab:stability_complete}
\end{table*}

\subsubsection{Impact of Verbalizer designs} 
During self-supervised tuning, the model saw a large number of first sentences as options, which may contain similar options to the unseen tasks, thus it may have better generalization capabilities. To test how robust the model is to the verbalizer changes compared with UniMC, we design 10 sets of verbalizers for SST-2 and IMDb, covering various scenarios: 1) verbalizers with a single word; 2) verbalizers with different punctuation marks; 3) combinations of single verbalizers; 4) different format for different classes. 
% Humans can easily differentiate which verbalizer is negative and which is positive. 
For a fair comparison, we only use one of our checkpoints and compare it with the UniMC checkpoint released. The results are shown in Table \ref{tab:stability_complete}. 
% The result shows that SSTuning-ALBERT performs better than UniMC on average. At the same time, SSTuning-ALBERT also has more stable performances over different verbalizers. 
We find that SSTuning-ALBERT performs better on average and is more stable. For the most challenging case, which is \textit{"Terrible!"} and \textit{"I like the movie! It is wonderful!"}, SSTuning-ALBERT outperforms UniMC by 20.4 points for SST-2 and 17 points for IMDb.

\subsubsection{Classification Mechanism} \label{sec:mechanism}
\begin{figure}[t]
    \centering
    \includegraphics[width=0.6\linewidth]{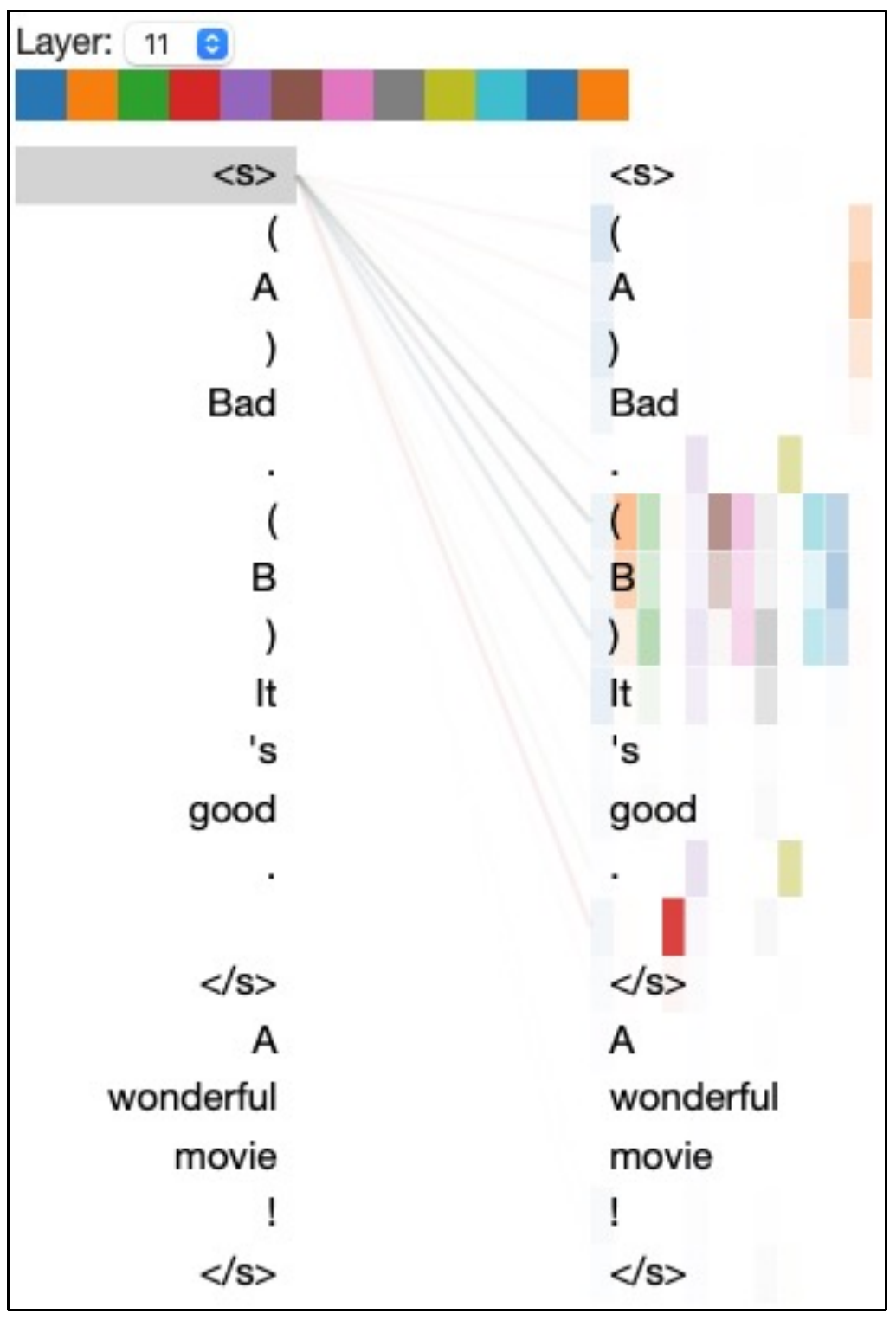}
    \caption{Attention map of \texttt{[CLS]} token (which is \texttt{<s>} here for RoBERTa backbone) in the last layer for a movie review. This figure is generated with BertViz \citep{acl19/bertviz}.}
    \label{fig:attention_map}
\end{figure}

To investigate how our models make correct decisions, we did a case study on a movie review example. As shown in Figure \ref{fig:attention_map}, we used SSTuning-base (number of labels configured as 2) to classify whether the movie review "A wonderful movie!" is negative or positive. We set the verbalizers as "Bad." and "It's good." to see how the length of options impacts the decision. The prediction of the model is 1, which is correct. We find that \texttt{[CLS]} token attends more to the second opinion, especially to the tokens around the index indicator "B" in the last layer. This is consistent with our intuitions. For humans, when we do classification tasks, we normally compare the options and select the option that best matches the text. We show additional attention maps and analysis in Appendix \ref{append:mechanism}.

\subsubsection{Importance of Index Indicators}
% As our usage of the [CLS] token is unconventional, it may be hard to understand the working mechanism of the model. Normally the output of the transformer encoder for text classification is the embedding of the text and the output layer contains the embedding for each class. Intuitively, the [CLS] token for our model is the embedding of the correct label index. To further understand the information flow, we did the following experiment. 

% To understand how important the index indicator is for the model to make correct predictions, we did the following experiment.
% Firstly, we apply different formats to the index indicator, which are: 1) alphabet characters in upper case (A, B, C...), which is the default format; 2) numerical index (0, 1, 2...); 3) same index indicator (0, 0, 0...). Then we tune SSTuning-base with the settings. For zero-shot inference, we use the same index indicator as tuning. We also tried changing the index indicators that are inconsistent with tuning to see the impact. The results are shown in Table \ref{tab:index_indicator}. 

% \isak{
To further understand how the index indicator guides the model to make the prediction, we employ different indicator designs during the tuning and inference stage. Specifically, we consider different formats of the index indicator, which are: 1) alphabet characters (A, B, C...), which is the default format; 2) numerical index (0, 1, 2...); 3) same index indicator for all options (0, 0, 0...). During the inference, we also consider two special indicators: 4) same alphabet characters (A, A, A...), and 5) rearranged alphabet characters (B, A, D, C...). The results are shown in Table \ref{tab:index_indicator}. There is not much difference between using alphabet characters and numerical indexes, as shown in cases 1 and 2. As shown in case 3, using the same characters will degrade the performance but not much, which means the model can rely on position embedding of the index indicator to make the correct predictions. As shown in cases 4 and 5, using inconsistent index indicators will greatly degrade the performance, which further verifies the importance of using consistent index indicators to make correct predictions.
% Then discuss your findings and conclusions...}

\begin{table}[]
    \centering
    \small
    % \resizebox{0.4\textwidth}{!}{
    \begin{tabular}{lllcc}
    \toprule
         & \textbf{Tuning} & \textbf{Inference} & \textbf{Avg} & \textbf{Std} \\ 
         \midrule
         1& (A, B, C...) & (A, B, C...) & 75.9 & 0.3 \\
         2&(0, 1, 2...) & (0, 1, 2...) & 75.6 & 0.4 \\
         3&(0, 0, 0...) & (0, 0, 0...) & 74.1 & 0.6 \\
    \midrule
         4& (A, B, C...) & (A, A, A...) & 32.0 & 1.1\\
         5&(A, B, C...) & (B, A, D, C...) & 23.4 & 12.1\\
    \bottomrule
    \hline
    \end{tabular}%}
    \caption{Performance with same and different index indicators during tuning and inference. “Std” indicates Standard Deviation.
    % \isak{no need to show std (not so important here), can use the space to show more complete prompt design}
    }
    \label{tab:index_indicator}
\end{table}

\subsubsection{Impact of Hard Negative Samples}
\begin{figure}[t]
    \centering
    \includegraphics[width=0.6\linewidth]{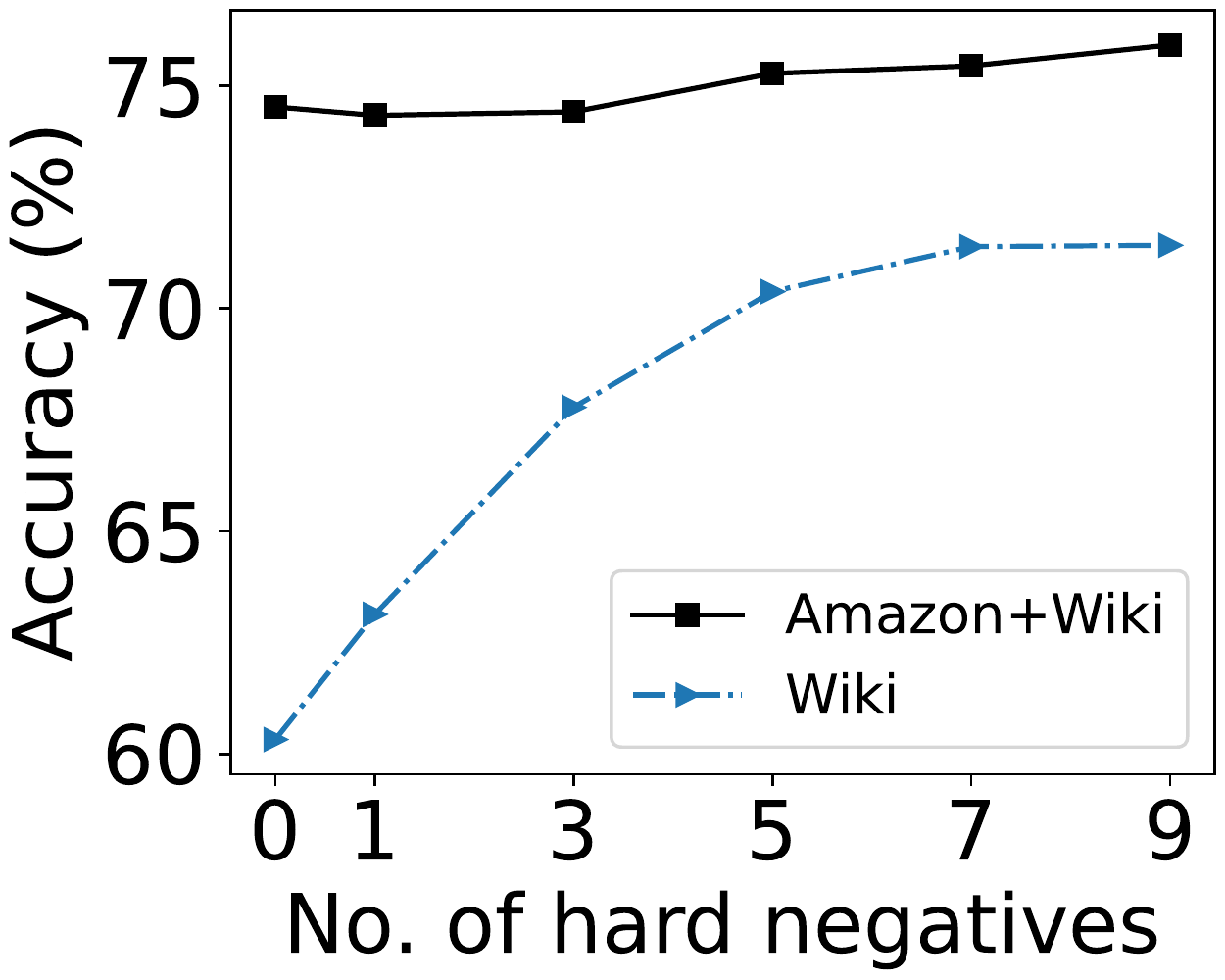}
    \caption{Zero-shot accuracy with different numbers of hard negatives.}
    \label{fig:n_hard}
\end{figure}

Intuitively, adding more hard negatives will make the task more difficult, thus forcing the mode to better understand the semantics of the sentences. We tested the impact of hard negatives based on two settings: 1) train with both the Amazon reviews and Wikipedia, each with 2.56M samples; 2) train with only 2.56M Wikipedia samples. We don't train with only Amazon reviews since they don't have hard negatives. %\isak{briefly explain why these two (readers may not remember you don't have negative samples for Amazon)} 
The results with $0, 1, 3, 5, 7, 9$ hard negatives are shown in Figure 
% \ref{fig:n_hard_sample}(a).
\ref{fig:n_hard}. 

In general, adding more hard negatives will improve the performance. For the case with both datasets, the impact of hard negatives is small. This is because the Amazon review dataset alone can achieve good performance, as shown in Table \ref{tab:impact_dataset}. However, hard negatives have a significant impact on the setting with only Wikipedia for tuning. The possible reason is that without hard negatives the model may only learn keyword matching instead of semantics since the keywords may appear many times in the same Wikipedia article. %\isak{the last sentence is unclear}

% \begin{figure}[t]
%     \centering
%     \includegraphics[width=0.8\linewidth]{figs/n_hard.pdf}
%     \caption{Zero-shot accuracy with different numbers of hard negatives. Mean accuracies of all the datasets over 5 seeds are reported.}
%     \label{fig:n_hard}
% \end{figure}

\subsubsection{Additional Analysis}
We report additional analysis in Appendix \ref{sec:addional_results}. As shown in Figure \ref{fig:n_sample}, we can further improve the performance by increasing the tuning sample size. 
% Following the evaluation method in \citep{emnlp22/uniMC}, we analyze the impact of verbalizer designs in SST-2 and IMDb tasks. As shown in Table \ref{tab:stability_complete}, compared to UniMC, SSTuning-ALBERT is more robust to verbalizer variations. 
We also compared SSTuning-base with different numbers of output labels $N_\text{model}$. As shown in Appendix \ref{append:label_num}, we can increase $N_\text{model}$ to inference on datasets with more classes.

\section{Related Work}
% \isak{better keep the related work within a half page, currently too long}
\paragraph{Zero-shot text classification.}
% \subsection{Zero-Shot Text Classification}
% Compared with the "pre-training and fine-tuning paradigm" for language models \citep{naacl19/bert,GPT1,jmlr20/T5,acl20/bart}, 
Zero-shot learning has the advantage that no annotated data is required for downstream tasks.
% , but is much more challenging. \isak{cannot compare method v.s. setting, can delete and directly introduce ZSL}
Prompting-based methods \citep{npls20/gpt3,PaLM,NAACL21/iPET,acl21/lm-bff} that reformulate the inputs as prompts can perform much worse in the zero-shot setting than few-shot settings as it may be hard for the PLMs to interpret the templates. A better option may be mining-based method \citep{emnlp22/mining-based}, which mines the labeled data from the unlabeled corpus for fine-tuning each downstream task. Similarly, generation-based approaches \citep{nlps22/superGen,corr/zerogen} generate labeled data with a generative PLM. %The two types of methods need to fine-tune the model for each task, which hinders their usage. 

More works on zero-shot text classifications are based on transfer learning. Instruction-tuning-based models like FLAN \citep{iclr22/flan} and T0 \citep{iclr22/T0}, fine-tine PLMs on a collection of datasets described by instructions or prompts to improve performances on unseen tasks.
% UnifiedQA \citep{emnlp20/unifiedQA} formats multiple tasks as the question-answering format. After fine-tinging on a collection of tasks, the model can perform well on unseen tasks. 
PLMs can also be meta-tuned \citep{emnlp21/meta-tuning} on text classification datasets and do zero-shot on other classification datasets.
UniMC \citep{emnlp22/uniMC} converts several tasks to multiple-choice tasks and does zero-shot inference on tasks that can be formulated in the same format. Another line of work is to convert text classification problems to textual entailment problems. By fine-tuning on natural language inference datasets \citep{emnlp19/TE} or a dataset from Wikipedia \citep{naacl22/TE_TowardsOT}, the models can do inference directly on text classification datasets. 
% Our approach combines the advantages of the previous methods. 
Instead of using annotated datasets, we only need unlabeled data to generate a large number of labeled samples as tuning and validation sets by exploring the inherent text structure. 
% After tuning, the model can be used for various downstream classification tasks without further adjustment. 
% \isak{can delete this paragraph}

\paragraph{Self-supervised learning.} 
% \subsection{Self-Supervised Learning} 
Self-supervised learning has been widely applied during language model pre-training by leveraging the input data itself as supervision signals \citep{tkde/ssl_generative_or_contrastive}. Left-to-right language modeling \citep{GPT1} and masked language modeling \citep{naacl19/bert,corr/roberta,iclr20/albert} help learn good sentence representations. In order to capture the sentence-level relations of downstream tasks, \citet{naacl19/bert} pre-train a next sentence prediction task and
% , which is to predict whether $S_2$ is the next sentence that follows $S_1$, given a sentence pair ($S_1$, $S_2$) as input. \isak{no special marker or notation in related work} 
\citet{iclr20/albert} use sentence order prediction task to model the inter-sentence coherence. \citet{iclr20/StructBERT} combine the two objectives to form a three-way classification task.
% , which can predict whether $S_2$ is a sentence that follows $S_1$, a sentence that precedes $S_1$, or a sentence randomly sampled from another document. 
Instead of modeling the inter-sentence relations, \citet{nips21/coco-lm} employs sequence contrastive learning to align the corrupted text sequences that originate from the same input source and guarantee the uniformity of the representation space. %\isak{no need to use special font in related work}
% \isak{no our work, can delete the following sentence} 
Our work uses a harder learning objective called first sentence prediction: given several options and text, find the corresponding first sentence preceding the text. 
% In order to solve this task, the model needs to identify each option and text, and then figure out the relationship between the text and each option. The results of the evaluation datasets show that the model can accomplish this task well by seeing a large number of examples. 

\section{Conclusions}

In this work, we propose a new learning paradigm called SSTuning for zero-shot text classification tasks. By forcing the model to predict the first sentence of a paragraph given the rest, the model learns to associate the text with its label for text classification tasks. Experimental results show that our proposed method outperforms state-of-the-art baselines on 7 out of 10 tasks and the performance is more stable with different verbalizer designs.
%[the following can be deleted depending on the length, you can even add back on camera-ready version]
Our work proves that applying self-supervised learning is a promising direction for zero-shot learning. In the future, we plan to apply SSTuing to other tasks by designing proper learning objectives.

% In this work, we proposed \isak{propose} \isak{same tense in the whole article} a new learning paradigm called SSTuning for zero-shot classification tasks. Together with the new learning objective called FSP and a new data generation method, the models obtain good zero-shot classification capability by tuning on unlabeled data. Without further training, the tuned model outperforms state-of-the-art baselines on 7 out of 10 tasks. Our work proves that applying self-supervised learning at tuning stage is a promising direction for zero-shot learning. In addition, since FSP only need unlabeled data, it has the potential to be applied to language model pre-training. In the future, we plan to apply SSTuing to other tasks by designing proper learning obejectives.

\section*{Limitations}
In this work, we proposed SSTuning for zero-shot text classification tasks. During inference, we may need to design verbalizers even though we can use templates like "This
text is about [label name]". For simplicity and fair comparison, we only refer to previous works for such designs, which may be sub-optimal. As shown in Table \ref{tab:stability_complete}, using the verbalizers "Terrible." and "Great." work better than "It's terrible." and "It's great." for the SST-2 and IMDA tasks that we reported in the main results. If the labeled validation set is provided, the model may perform better by choosing verbalizers based on the validation set.

Due to limited computation resources, we only tuned the model with 5.12 million samples, which is only a small portion of the available samples. We believe that tuning the model on a larger dataset help improve the performance. Even though the computational cost will also increase, it is worth it since no more training is needed at the inference phase. In addition, we did not do extensive hyperparameter searches except for the learning rate, which may further improve the performance. 

In our experiment, we only tested the method with discriminative models like RoBERTa and ALBERT. Its performance with generative models is not known. It is non-trivial to test on such models since generative models can do both natural language understanding tasks and natural language generation tasks. We leave this as future work.

% \section*{Ethics Statement}
% Ethics Statement

\section*{Acknowledgements}
% Acknowledgements
% This research is supported, in part, by Alibaba Group through Alibaba Innovative Research (AIR) Program and Alibaba-NTU Singapore Joint Research Institute (JRI) (Alibaba-NTU-AIR2021B6), Nanyang Technological University, Singapore. This research is also supported by the Ministry of Education Tier 1 grant (MOE Tier 1 RS21/20).

This research is supported, in part, by Alibaba Group through Alibaba Innovative Research (AIR) Program and Alibaba-NTU Singapore Joint Research Institute (JRI), Nanyang Technological University, Singapore. Chaoqun Liu and Guizhen Chen extend their gratitude to Interdisciplinary Graduate Programme and School of Computer Science and Engineering, Nanyang Technological University, Singapore, for their support. This research is also supported by the Ministry of Education Tier 1 grant (MOE Tier 1 RS21/20).

% Entries for the entire Anthology, followed by custom entries
\bibliography{anthology,custom}
\bibliographystyle{acl_natbib}

\appendix
% \clearpage
\appendix
\section{Additional Dataset Details} \label{sec:appendix}

\subsection{Tuning Datasets} 
\label{sub_sec:data_filter}
The original unlabeled datasets can be noisy and some paragraphs are not suitable for generating tuning datasets. We filter the paragraphs with the following features: 
1) the paragraph only contains 1 sentence; 2) the first sentence contains less than or equal to 3 characters; 3) the first sentence only contains non-alphabetic symbols; 4) repeated paragraphs. Some of the final generated samples from English Wikipedia and Amazon product reviews are shown in Table \ref{tab:example_tuning}.

\subsection{Evaluation Datasets}
We summarize the dataset statistics for the evaluation datasets in Table \ref{tab:stats}. We download all the datasets from Huggingface \citep{emnlp21/huggingface_datasets}, except 20newsgroup. For Yahoo Topics, we concatenate the question and answer as inputs. For DBPedia and Amazon, we concatenate the title and content. For 20newsgroup, we follow the recommendations to remove headers, footers, and quotas\footnote{\url{https://scikit-learn.org/0.19/datasets/twenty\_newsgroups.html}}. However, if the text becomes empty after removing the components, we will use the original text instead.

The verbalizers for each dataset are shown in Table \ref{tab:verbalizer}. We try to unify the verbalizer design for similar tasks. For topic classification tasks, we use the template \textit{"This text is about []."} after converting the class names to meaningful words. For binary classifications, we use \textit{"It's terrible."} for negative class and \textit{"It's great."} for positive class. For SST-5, we refer to \citep{acl21/lm-bff} to design the verbalizers. Some of the reformulated text for the evaluation datasets are shown in Table \ref{tab:example_evaluation}.

\begin{table}[]
\centering
\small
\begin{tabular}{l c c c c}
    \toprule
    \textbf{Dataset} & \textbf{\# Class} & \textbf{\# Train} & \textbf{\# Val} & \textbf{\# Test} \\
    \midrule
    % \multirow{4}*{}
    Yahoo. & 10 & 1.4M  & 0 & 60k\\
    AG News & 4 & 120k  & 0 & 7.6k\\
    DBPedia & 14 & 560k  & 0 & 70k\\
    20 News. & 20 & 11,314 & 0 & 7532\\
    \midrule
    % \multirow{6}*{}
    SST-2 & 2 & 67,349  & 872 & 0 \\
    IMDB & 2 & 25k  & 0 & 25k\\
    Yelp & 2 & 560k  & 0 & 38k\\
    MR & 2 & 8,530 & 1,066 & 1,066 \\
    Amazon & 2 & 3.6M & 0 & 400k \\
        SST-5 & 5 & 8,544  & 1,101 & 2,210\\
    \bottomrule
\end{tabular}
\caption{Dataset statistics for evaluation datasets}
\label{tab:stats}
\end{table}

\section{Additional Experiment Details}
\subsection{Experiment setup}
The hyperparameters for the main results (Section \ref{sec:main_results}) are shown in Table \ref{tab:hyperparameters}. We try to use the same settings as much as possible. The training time for the three SSTuning models is with 5.12M tuning samples and 64k validation samples (also generated via FSP).

\subsection{Additional Results} \label{sec:addional_results}
% \subsubsection{How does the number of training samples impact?}
% \paragraph{Impact of Tuning Sample Size.} 
\subsubsection{Impact of Tuning Sample Size} 
\begin{figure}[t]
    \centering
    \includegraphics[width=0.8\linewidth]{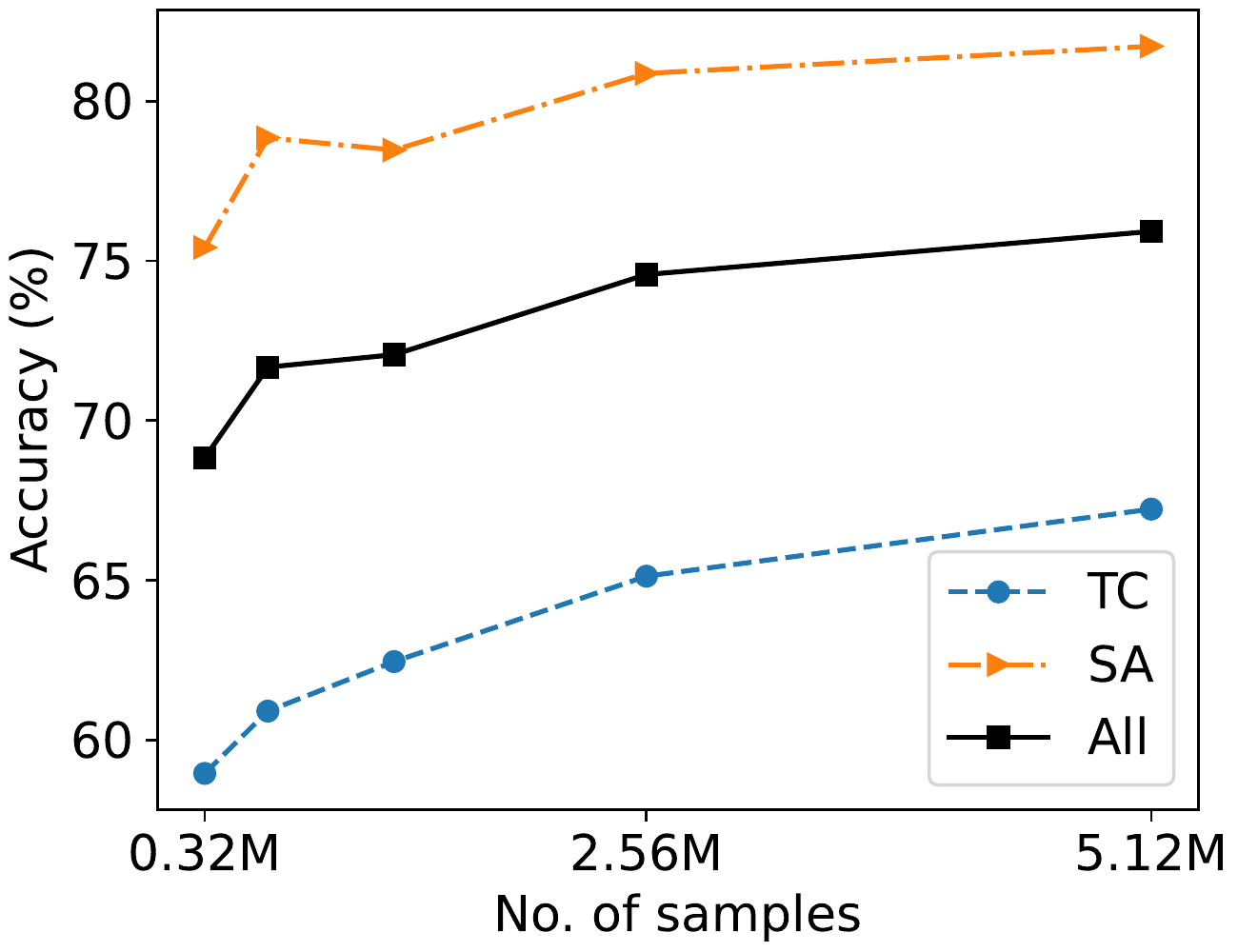}
    \caption{Zero-shot accuracy with different training sample sizes. Mean accuracy over 4 topic classification tasks, 6 sentiment analysis tasks, and all the tasks are reported.}
    \label{fig:n_sample}
\end{figure}

To test how the tuning sample size impacts the performance, we trained SSTuning-base with 320k, 640k, 1.28M, 2.56M, and 5.12M samples, with half generated from Wikipedia and half from Amazon reviews. The results are shown in Figure 
% \ref{fig:n_hard_sample}(b). 
\ref{fig:n_sample}.
With more samples, the performances are increasing in general, especially for topic classification tasks. With such observation, it is likely to further improve the performance by increasing the tuning sample size. Even though tuning on larger datasets is more computationally expensive, it is worth doing since no further training is required for downstream tasks.

% \begin{figure}[t]
%     \centering
%     \includegraphics[width=0.6\linewidth]{figs/n_sample_v2_big.pdf}
%     \caption{Zero-shot accuracy with different training sample sizes. }
%     \label{fig:n_sample}
% \end{figure}

% % \paragraph{Impact of Verbalizer designs.} 
% \subsubsection{Impact of Verbalizer designs} 
% During self-supervised tuning, the model saw a large number of first sentences as options, which may contain similar options to the unseen tasks, thus it may have better generalization capabilities. To test how robust the model is to the verbalizer changes compared with UniMC, we design 10 sets of verbalizers for SST-2 and IMDb, covering various scenarios: 1) verbalizers with a single word; 2) verbalizers with different punctuation marks; 3) combinations of single verbalizers; 4) different format for different classes. 
% % Humans can easily differentiate which verbalizer is negative and which is positive. 
% For a fair comparison, we only use one of our checkpoints and compare it with the UniMC checkpoint released. The results are shown in Table \ref{tab:stability_complete}. 
% % The result shows that SSTuning-ALBERT performs better than UniMC on average. At the same time, SSTuning-ALBERT also has more stable performances over different verbalizers. 
% We find that SSTuning-ALBERT performs better on average and is more stable. For the most challenging case, which is \textit{"Terrible!"} and \textit{"I like the movie! It is wonderful!"}, SSTuning-ALBERT outperforms UniMC by 20.4 points for SST-2 and 17 points for IMDb.

\subsubsection{Impact of the Number of Output Labels}  \label{append:label_num}
In our main results, we set the number of output labels $N_\text{model}$ as 20. However, a classification dataset may have more than 20 classes. To test the scalability of the label number, we tune another variant for SSTuning-base. We use numerical numbers (0, 1, 2...) as the index indicator and set $N_\text{model}$ as 40. The comparison between the two versions is shown in Table \ref{tab:result_label_num}. Increasing $N_\text{model}$ from 20 to 40 only degrade the performance by 1.4 points (75.9\% to 74.5\%), showing the good scalability of our approach. As an alternative for the datasets with more classes, we can split the labels and do a multi-stage inference.

% \paragraph{Attention Map.} \cq{The example in the appendix should have the same options as main text.}
\subsubsection{Classification Mechanism} \label{append:mechanism}
\begin{figure*}[h]
    \centering
    % \small
    % \includegraphics[width=1\linewidth]{figs/attention_map1.pdf}
    \includegraphics[width=1\linewidth]{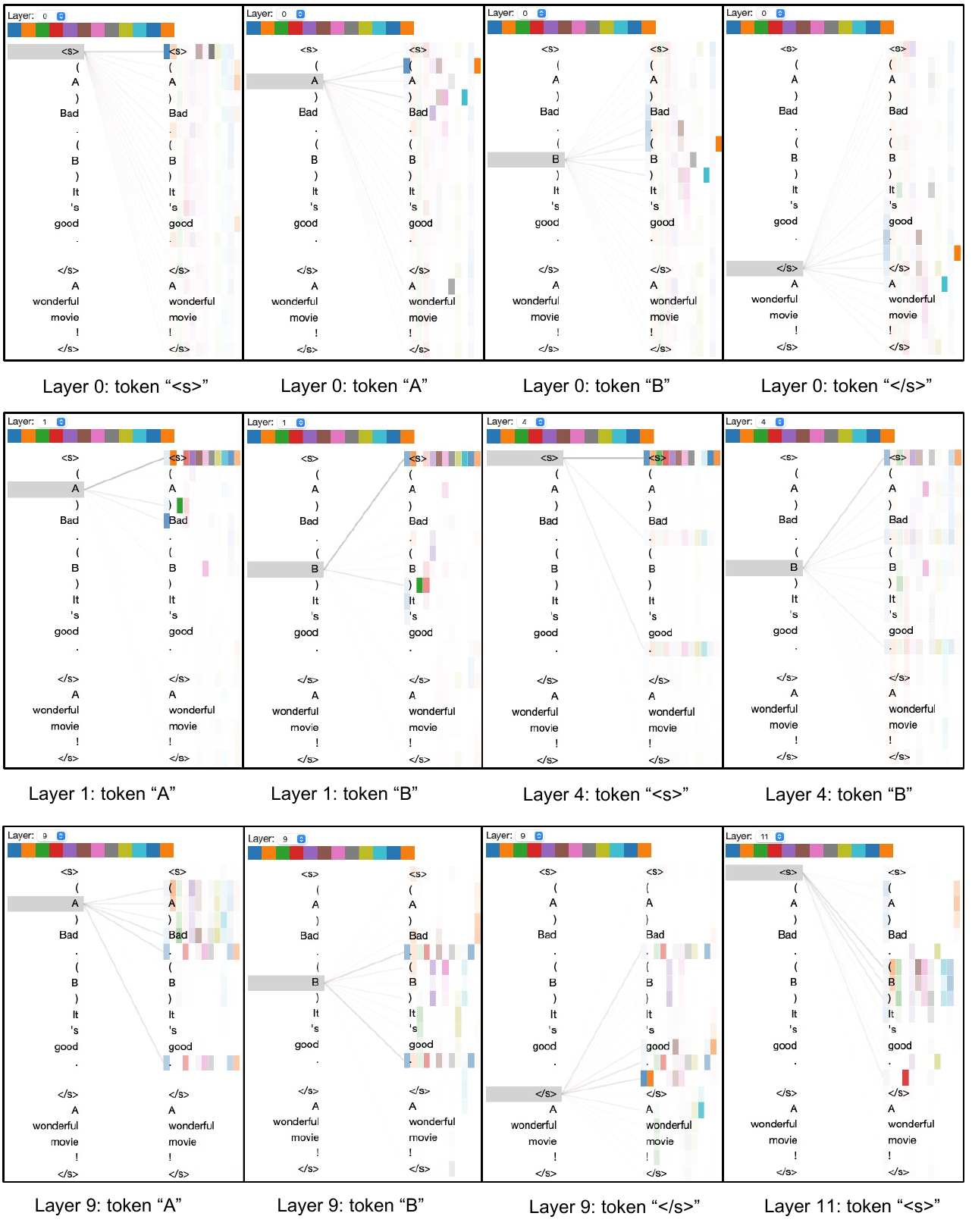}
    \caption{Attention map for a movie review example. The original text is "A wonderful movie!" and the verbalizers are "Bad." and "It's Good.". The model is SSTuning-base with 2 classes. }
    % The visualization tool for this example is BertViz \citep{acl19/bertviz}.}
    \label{fig:attention_map_full}
\end{figure*}

We plot more attention maps for the example discussed in Section \ref{sec:mechanism}  in Figure \ref{fig:attention_map_full}. We focus on a few important tokens, including the classification token \texttt{<s>}, the option indicators \texttt{A} and \texttt{B}, and the separator token \texttt{</s>}. In Layer 0, \texttt{<s>} attends to all the options and the text. \texttt{A} and \texttt{B} attend more to its own options. \texttt{</s>} attend more to the text tokens. 
In higher layers, \texttt{A} and \texttt{B} attend even more to their own option tokens (Layer 1) but also have some interactions (Layer 4). In layer 9, \texttt{A} and \texttt{B} attend more its own option tokens again and also the period mark, while \texttt{</s>} attend to both the text tokens and the options tokens for \texttt{B} (the positive option). In the end, \texttt{<s>} attends to \texttt{B}, which is the positive option. Based on the observations, we hypothesize that the model has the capability to encode the options and text separately, compare the options and text, and choose the positive option in the end.

\begin{table*}[t]
\centering
\small
\begin{tabular}{C{0.15\textwidth}p{0.8\textwidth}}
    \toprule
    \textbf{Dataset} & \textbf{Verbalizers} \\
    \midrule
    Yahoo Topics & "This text is about society \& culture.", "This text is about science \& mathematics.", "This text is about health.", "This text is about education \& reference.", "This text is about computers \& internet.", "This text is about sports.", "This text is about business \& finance.", "This text is about entertainment \& music.", "This text is about family \& relationships.", "This text is about politics \& government." \\
    \midrule
    AG News & "This text is about politics.", "This text is about sports.", "This text is about business.", "This text is about technology."\\
    \midrule
    DBPedia &  "This text is about company.", "This text is about educational institution.", "This text is about artist.", "This text is about athlete.", "This text is about office holder.", "This text is about mean of transportation.", "This text is about building.", "This text is about natural place.", "This text is about village.", "This text is about animal.", "This text is about plant.", "This text is about album.", "This text is about film.", "This text is about written work."\\ 
    \midrule
    20 Newsgroup & "This text is about atheism.", "This text is about computer graphics.", "This text is about microsoft windows.", "This text is about pc hardware.", "This text is about mac hardware.", "This text is about windows x.", "This text is about for sale.", "This text is about cars.", "This text is about motorcycles.", "This text is about baseball.", "This text is about hockey.", "This text is about cryptography.", "This text is about electronics.", "This text is about medicine.", "This text is about space.", "This text is about christianity.", "This text is about guns.", "This text is about middle east.", "This text is about politics.", "This text is about religion."\\
    \midrule
    SST-2, IMDB, Yelp, MR, Amazon & "It's terrible.", "It's great."\\
    % \midrule
    % SST-2 & "It's terrible.", "It's great."\\
    % \midrule 
    % IMDB & "It's terrible.", "It's great."\\
    % \midrule 
    % Yelp & "It's terrible.", "It's great."\\
    % \midrule 
    % MR & "It's terrible.", "It's great."\\
    % \midrule 
    % Amazon& "It's terrible.", "It's great."\\
    \midrule 
    SST-5 & "It's terrible.", "It's bad.", "It's okay.", "It's good.", "It's great." \\
    \bottomrule
\end{tabular}
\caption{Verbalizers for the evaluation datasets.}
\label{tab:verbalizer}
\end{table*}

\begin{table*}[h]
\centering
\small
\begin{tabular}{l c c c }
    \toprule
    \textbf{Parameter} & \textbf{Fine-tuning} & \textbf{SSTuning-base/SSTuning-large} & \textbf{SSTuning-ALBERT}\\
    \midrule
     Model & RoBERTa\textsubscript{\text{large}} (355M) & RoBERTa\textsubscript{\text{base}}/RoBERTa\textsubscript{\text{large}} (355M) & ALBERT\textsubscript{\text{xxlarge}}(V2)(235M)\\
     Model Selection & Best & Best & Best \\
     Batch Size & 16 & 128  & 64\\
     Precision & FP16 & FP16 & FP16\\
     Optimiser & AdamW & AdamW & AdamW\\
     Learning Rate & 1e-5 & 2e-5 & 1e-5\\
     LR Scheduler & linear decay & linear decay & linear decay \\
     AdamW Epsilon & 1e-8  & 1e-8 & 1e-8\\
     AdamW $\beta_1$ & 0.9 & 0.9 & 0.9\\
     AdamW $\beta_1$ & 0.999  & 0.999 & 0.999\\
     Weight Decay & 0.01 & 0.01 & 0.01\\
     Classifier Dropout & 0.1  & 0.1 & 0.1\\
     Attention Dropout & 0.1 & 0.1 & 0\\
     Hidden Dropout & 0.1 & 0.1 & 0\\
     Max Steps & - & 40000 & 80000\\
     Max Epochs & 3 & 1 & 1\\ \midrule
     Hardware &  1 NVIDIA V100 & 8 NVIDIA V100 &  4 NVIDIA A100 \\
 Training time & - & 3h/8h & 31h\\
     % Batch Sampler & - \\
     \bottomrule
\end{tabular}
\caption{Hyperparameters and training information for full-shot fine-tuning, SSTuing-base, SSTuning-large and SSTuing-ALBERT.}
\label{tab:hyperparameters}
\end{table*}

\begin{table*}[h]
    \centering
    \small
    \begin{tabular}{C{0.09\textwidth}C{0.06\textwidth}C{0.13\textwidth}p{0.6\textwidth}}
    \midrule
        \textbf{Dataset} & \textbf{Label} & \textbf{Positive Option} & \textbf{Generated Text} \\ \midrule
        Wikipedia & 12 (M) & In parliament, Satouri serves on the Committee on Employment and Social Affairs and the Subcommittee on Security and Defence. & (A) [PAD] (B) The work of lojas, are found in both the town and the countryside. (C) [PAD] (D) [PAD] (E) [PAD] (F) [PAD] (G) In 1848 riots and looting took place, and in 1849 an epidemic broke out. (H) [PAD] (I) Leptostylus retrorsus is a species of beetle in the family Cerambycidae. (J) The 2020 – 21 Russian Football National League was the 29th season of Russia's second - tier football league since the dissolution of the Soviet Union. (K) [PAD] (L) He opposed several times to the decisions of his party, as when Congress was dissolved in 2019, he supported Martín Vizcarra's measure and did not attend to the inauguration of Vice President Mercedes Araoz. (M) In parliament, Satouri serves on the Committee on Employment and Social Affairs and the Subcommittee on Security and Defence. (N) [PAD] (O) [PAD] (P) [PAD] (Q) [PAD] (R) [PAD] (S) The church has a rectangular nave with stone walls that are around 2 meters thick. (T) On February 2,, the Blue Jays and Downs agreed to a one - year, \$ 1. 025 million contract, avoiding the arbitration process. [SEP] In addition to his committee assignments, he is part of the parliament's delegations to the Parliamentary Assembly of the Union for the Mediterranean and for relations with the NATO Parliamentary Assembly. \\ \midrule
        Wikipedia & 0  (A) & Rawat emigrated to Canada from India in 1968. & (A) Rawat emigrated to Canada from India in 1968. (B) Meskowski was a racing car constructor. (C) [PAD] (D) , there were 42 people who were single and never married in the municipality. (E) [PAD] (F) [PAD] (G) [PAD] (H) [PAD] (I) [PAD] (J) [PAD] (K) It is a Church of England school within the Diocese of Salisbury. (L) Falkoner Allé was opened to the public after Hømarken ( literally " Hayfield " ), an area to the north belonging to Ladegården, originally a farm under Copenhagen Castle, was auctioned off. (M) [PAD] (N) [PAD] (O) In the fall of her senior year at McDonogh, Cummings committed to play for the University of Maryland\'s women\'s lacrosse team as the nation\'s top recruit. (P) Ranville is a native of Flint, Michigan and attended St. Agnes High School. (Q) The Dodge\'s Institute of Telegraphy was housed in the Institutes building at 89 East Monroe. (R) During 2004 - 2011, Rawat was President of the Communications Research Centre, Canada\'s centre of excellence for telecommunications R \& D, with 400 staff and an annual budget of over \$ 50 million. (S) [PAD] (T) [PAD] [SEP] She speaks English, French, Hindi and Spanish. \\ \midrule
        Amazon Product Review & 1 (B) & This popcorn is really best suited for kettle corn. & (A) [PAD] (B) This popcorn is really best suited for kettle corn. (C) Professional Quality with Amazing results. (D) [PAD] (E) [PAD] (F) [PAD] (G) I found my new S6 to be a little TOO thin, and so slick it was sliding off of everything, so I wanted a clear bumper. (H) Excellent price. (I) [PAD] (J) [PAD] (K) I\'ve always loved Bounce dryer sheets, but was not too fond of the synthetic " Outdoor Fresh " scents. (L) [PAD] (M) [PAD] (N) [PAD] (O) I cut the cord and bought this mohu leaf antenna to get the local channels. (P) [PAD] (Q) The product came pretty quickly with very easy instructions. (R) [PAD] (S) [PAD] (T) Watch Land Before Time and had to have one for Xmas. [SEP] The kernels pop up to a nice large size. Don\'t think I would compare them to mushrooms - button mushrooms maybe (LOL). They are a bit on the chewy side if you go the butter route. They are really best as crisp, salty-sweet kettle corn. Yum! We use a Whirley Pop for popcorn--our favorite kitchen "appliance"! Don\'t know if some other method would make the popcorn crisper. No matter--would buy this again just for the way it tastes as kettle corn! \\ \midrule
        Amazon Product Review & 18 (S) & Works pretty good. & (A) [PAD] (B) [PAD] (C) [PAD] (D) [PAD] (E) [PAD] (F) [PAD] (G) [PAD] (H) [PAD] (I) [PAD] (J) [PAD] (K) [PAD] (L) [PAD] (M) [PAD] (N) [PAD] (O) [PAD] (P) Great value for a creeper. (Q) [PAD] (R) [PAD] (S) Works pretty good. (T) [PAD] [SEP] Just wish the fm stations on the device would go lower. The best one in my area is 85.1 but the device only goes to 88.1. Still a great product. \\ \midrule
        
    \end{tabular}
    \caption{Examples generated for SSTuning with English Wikipedia and Amazon product review dataset.}
    \label{tab:example_tuning}
\end{table*}

\begin{table*}[h]
    \centering
    \small
    \begin{tabular}{C{0.09\textwidth}C{0.05\textwidth}C{0.13\textwidth}p{0.6\textwidth}}
    \midrule
        \textbf{Dataset} & \textbf{Label} & \textbf{Positive Option} & \textbf{Reformulated Text} \\ \midrule
        % Yahoo Topics & 5 & This text is about sports. & (A) This text is about society \& culture. (B) This text is about science \& mathematics. (C) This text is about health. (D) This text is about education \& reference. (E) This text is about computers \& internet. (F) This text is about sports. (G) This text is about business \& finance. (H) This text is about entertainment \& music. (I) This text is about family \& relationships. (J) This text is about politics \& government. (K) [PAD] (L) [PAD] (M) [PAD] (N) [PAD] (O) [PAD] (P) [PAD] (Q) [PAD] (R) [PAD] (S) [PAD] (T) [PAD] [SEP] Do you think Manchester United will catch Chelsea for the few remaining games?  No.  United blew their chance by drawing with Sunderland.  Chelsea now only need one point from 3 matches to win the Premiership.  United will beat Chelsea at Stamford Bridge to keep the race going a little longer though \\ \midrule
        AG News & 3 (D) & This text is about technology. & (A) This text is about politics. (B) This text is about sports. (C) This text is about business. (D) This text is about technology. (E) [PAD] (F) [PAD] (G) [PAD] (H) [PAD] (I) [PAD] (J) [PAD] (K) [PAD] (L) [PAD] (M) [PAD] (N) [PAD] (O) [PAD] (P) [PAD] (Q) [PAD] (R) [PAD] (S) [PAD] (T) [PAD] [SEP] REVIEW: 'Half-Life 2' a Tech Masterpiece (AP) AP - It's been six years since Valve Corp. perfected the first-person shooter with "Half-Life." Video games have come a long way since, with better graphics and more options than ever. Still, relatively few games have mustered this one's memorable characters and original science fiction story. \\ \midrule
        DBPedia & 9 (J) & This text is about animal. & (A) This text is about company. (B) This text is about educational institution. (C) This text is about artist. (D) This text is about athlete. (E) This text is about office holder. (F) This text is about mean of transportation. (G) This text is about building. (H) This text is about natural place. (I) This text is about village. (J) This text is about animal. (K) This text is about plant. (L) This text is about album. (M) This text is about film. (N) This text is about written work. (O) [PAD] (P) [PAD] (Q) [PAD] (R) [PAD] (S) [PAD] (T) [PAD] [SEP] Periscepsia handlirschi.  Periscepsia handlirschi is a species of fly in the family Tachinidae. \\ \midrule
%         20 Newsgroup & 14 & This text is about technology. & (A) This text is about atheism. (B) This text is about computer graphics. (C) This text is about microsoft windows. (D) This text is about pc hardware. (E) This text is about mac hardware. (F) This text is about windows x. (G) This text is about for sale. (H) This text is about cars. (I) This text is about motorcycles. (J) This text is about baseball. (K) This text is about hockey. (L) This text is about cryptography. (M) This text is about electronics. (N) This text is about medicine. (O) This text is about space. (P) This text is about christianity. (Q) This text is about guns. (R) This text is about middle east. (S) This text is about politics. (T) This text is about religion. [SEP] 
% Robert McElwaine is the authoritative source of scientific data on Internet.
% He can be reached alt.fan.mc-elwaine... Spiros \\ \midrule
        SST-2 & 1 (B) & It's great. & (A) It's terrible. (B) It's great. (C) [PAD] (D) [PAD] (E) [PAD] (F) [PAD] (G) [PAD] (H) [PAD] (I) [PAD] (J) [PAD] (K) [PAD] (L) [PAD] (M) [PAD] (N) [PAD] (O) [PAD] (P) [PAD] (Q) [PAD] (R) [PAD] (S) [PAD] (T) [PAD] [SEP] charles ' entertaining film chronicles seinfeld 's return to stand-up comedy after the wrap of his legendary sitcom , alongside wannabe comic adams ' attempts to get his shot at the big time . \\ \midrule
        SST-5 & 3 (D) & It's good. & (A) It's terrible. (B) It's bad. (C) It's okay. (D) It's good. (E) It's great. (F) [PAD] (G) [PAD] (H) [PAD] (I) [PAD] (J) [PAD] (K) [PAD] (L) [PAD] (M) [PAD] (N) [PAD] (O) [PAD] (P) [PAD] (Q) [PAD] (R) [PAD] (S) [PAD] (T) [PAD] [SEP] u.s. audiences may find -lrb- attal and gainsbourg 's -rrb- unfamiliar personas give the film an intimate and quaint reality that is a little closer to human nature than what hollywood typically concocts . \\ \midrule
    \end{tabular}
    \caption{Examples after reformulation for 4 evaluation datasets.}
    \label{tab:example_evaluation}
\end{table*}

\begin{table*}[!ht]
    \centering
    % \setlength\tabcolsep{1pt}
    % \small
    \resizebox{\textwidth}{!}{%
    \begin{tabular}{lcccccccccccccc}
    \toprule
        & \multirow{2}{*}{$N_{\textbf{model}}$} & \multicolumn{5}{c}{\textbf{Topic Classification}} &\multicolumn{7}{c}{\textbf{Sentiment Analysis}} & \multirow{2}{*}{\textbf{Avg}} \\
        % \cline{4-7} \cline{9-14}
        \cmidrule{3-6} \cmidrule{8-13}
         &  & \textbf{yah} & \textbf{agn} & \textbf{dbp} & \textbf{20n} & & \textbf{sst2} & \textbf{imd} & \textbf{ylp} & \textbf{mr} & \textbf{amz} & \textbf{sst5} & & \\ 
         \midrule
        SSTuning-base    &20 & 59.1 & 79.9 & 82.7 & 47.2 && 86.4 & 88.2 & 92.9 & 83.8 & 94.0 & 45.0 && 75.9 \\ 
                            % &&& ($\pm0.7$) & ($\pm2.5$)$ & ($\pm1.9$)$ & ($\pm1.4$)$ && ($\pm0.7$)$ & ($\pm1.3$)$ & ($\pm0.7$)$ & ($\pm1.1$)$ & ($\pm0.3$)$ & ($\pm0.6$)$ && ($\pm0.3$)$ \\ 
        SSTuning-base   &40 & 58.0 & 79.3 & 79.8 & 49.1 && 84.4 & 88.2 & 91.7 & 82.2 & 93.3 & 39.4 && 74.5 \\ 
                            % &&& 0.6 & 0.6 & 1.1 & 1.6 && 0.5 & 0.8 & 0.6 & 0.8 & 0.3 & 2.5 && 0.3 \\
    \bottomrule
    \end{tabular}
    }
    \caption{Accuracy over different number of labels $N_{\textbf{model}}$.}
    \label{tab:result_label_num}
\end{table*}

\end{document}